\documentclass[journal]{IEEEtran}%
\ifCLASSOPTIONcompsoc
  \usepackage[nocompress]{cite}
\else
  \usepackage{cite}
\fi
\ifCLASSINFOpdf
\else
\fi
\usepackage{tikz}
\usepackage{adjustbox}
\usepackage{algorithm}               % format of the algorithm
\usepackage{algorithmic}             % format of the algorithm
\usepackage{booktabs}
\usepackage{epstopdf}
\usepackage{enumerate}
\usepackage{pdfpages}
\usepackage{afterpage}
\usepackage{color, colortbl}
\usepackage{footmisc}
\usepackage{etoolbox}
\usepackage{tablefootnote}
\usepackage[flushleft]{threeparttable}
\usepackage{pifont}% http://ctan.org/pkg/pifont
\usetikzlibrary{patterns}
\usepackage{multirow,makecell}
\usepackage{url}
\usepackage{amsthm}
\theoremstyle{definition}

\usepackage{graphicx}
\graphicspath{{./figures/}}
\usepackage{amsmath}
\usepackage{subfig}
\usepackage{makecell}

\makeatletter
\newcommand{\printfnsymbol}[1]{%
	\textsuperscript{\@fnsymbol{#1}}%
}
\makeatother

\begin{document}
%
% paper title
% Titles are generally capitalized except for words such as a, an, and, as,
% at, but, by, for, in, nor, of, on, or, the, to and up, which are usually
% not capitalized unless they are the first or last word of the title.
% Linebreaks \\ can be used within to get better formatting as desired.
% Do not put math or special symbols in the title.
\title{FADNet++: Real-Time and Accurate Disparity Estimation with Configurable Networks}
%
%
% author names and IEEE memberships
% note positions of commas and nonbreaking spaces ( ~ ) LaTeX will not break
% a structure at a ~ so this keeps an author's name from being broken across
% two lines.
% use \thanks{} to gain access to the first footnote area
% a separate \thanks must be used for each paragraph as LaTeX2e's \thanks
% was not built to handle multiple paragraphs
%
%
%\IEEEcompsocitemizethanks is a special \thanks that produces the bulleted
% lists the Computer Society journals use for "first footnote" author
% affiliations. Use \IEEEcompsocthanksitem which works much like \item
% for each affiliation group. When not in compsoc mode,
% \IEEEcompsocitemizethanks becomes like \thanks and
% \IEEEcompsocthanksitem becomes a line break with idention. This
% facilitates dual compilation, although admittedly the differences in the
% desired content of \author between the different types of papers makes a
% one-size-fits-all approach a daunting prospect. For instance, compsoc 
% journal papers have the author affiliations above the "Manuscript
% received ..."  text while in non-compsoc journals this is reversed. Sigh.

\author{
Qiang~Wang, Shaohuai~Shi, Shizhen~Zheng, Kaiyong~Zhao, Xiaowen~Chu\printfnsymbol{1}
\IEEEcompsocitemizethanks{\IEEEcompsocthanksitem Qiang Wang, Shizhen Zheng, Kaiyong Zhao and Xiaowen Chu are with the Department
of Computer Science, Hong Kong Baptist University. \protect\\E-mail: \{qiangwang, szzheng, kyzhao, chxw\}@comp.hkbu.edu.hk} 
\IEEEcompsocitemizethanks{\IEEEcompsocthanksitem Shaohuai Shi is with the Department of Computer Science and Engineering, The Hong Kong University of Science and Technology. \protect\\E-mail: shaohuais@cse.ust.hk}
\IEEEcompsocitemizethanks{\IEEEcompsocthanksitem The first two authors contributed equally to this work. Asterisk indicates corresponding author.}
}
\IEEEtitleabstractindextext{%
\begin{abstract}
Deep neural networks (DNNs) have achieved great success in the area of computer vision. The disparity estimation problem tends to be addressed by DNNs which achieve much better prediction accuracy than traditional hand-crafted feature based methods. However, the existing DNNs hardly serve both efficient computation and rich expression capability, which makes them difficult for deployment in real-time and high-quality applications, especially on mobile devices. To this end, we propose an efficient, accurate, and configurable deep network for disparity estimation named FADNet++. Leveraging several liberal network design and training techniques, FADNet++ can boost its accuracy with a fast model inference speed for real-time applications. Besides, it enables users to easily configure different sizes of models for balancing accuracy and inference efficiency. We conduct extensive experiments to demonstrate the effectiveness of FADNet++ on both synthetic and realistic datasets among six GPU devices varying from server to mobile platforms. Experimental results show that FADNet++ and its variants achieve state-of-the-art prediction accuracy, and run at a significant order of magnitude faster speed than existing 3D models. With the constraint of running at above 15 frames per second (FPS) on a mobile GPU, FADNet++ achieves a new state-of-the-art result for the SceneFlow dataset.

\end{abstract}

% Note that keywords are not normally used for peerreview papers.
}

% make the title area
\maketitle

% To allow for easy dual compilation without having to reenter the
% abstract/keywords data, the \IEEEtitleabstractindextext text will
% not be used in maketitle, but will appear (i.e., to be "transported")
% here as \IEEEdisplaynontitleabstractindextext when the compsoc 
% or transmag modes are not selected <OR> if conference mode is selected 
% - because all conference papers position the abstract like regular
% papers do.
\IEEEdisplaynontitleabstractindextext
% \IEEEdisplaynontitleabstractindextext has no effect when using
% compsoc or transmag under a non-conference mode.

% For peer review papers, you can put extra information on the cover
% page as needed:
% \ifCLASSOPTIONpeerreview
% \begin{center} \bfseries EDICS Category: 3-BBND \end{center}
% \fi
%
% For peerreview papers, this IEEEtran command inserts a page break and
% creates the second title. It will be ignored for other modes.
% \IEEEpeerreviewmaketitle

\section{Introduction}\label{sec:introduction}

% \IEEEPARstart{I}{t}
%
% how to process such compute- and memory-intensive models on portable and lowpower devices remains a concern.
Disparity estimation (also referred to as stereo matching) is a classical and important problem in robotics and autonomous driving for 3D scene reconstruction \cite{linestereo2015,orbslam22017,smart_iotj2020}. While traditional methods based on hand-crafted feature extraction and matching cost aggregation such as Semi-Global Matching (SGM) \cite{hirschmuller2007stereo}) tend to fail on those textureless and repetitive regions in the images, recent advanced deep neural network (DNN) techniques surpass them with decent generalization and robustness to those challenging patches, and achieve state-of-the-art performance in many public datasets \cite{zagoruyko2015learning}\cite{zbontar2016stereo}\cite{flownet}\cite{mayer2016large}\cite{psmnet2018}\cite{ganet2019}. 
However, how to design an efficient DNN structure for disparity estimation with limited computational cost for those Internet-of-Things (IoT) scenarios remains a concern.

The DNN-based methods for disparity estimation are end-to-end frameworks which take stereo images (left and right) as input to the neural network and predict the disparity directly. The architectures of DNN are very essential to achieve accurate estimation, and can be categorized into two classes, the encoder-decoder network with 2D convolution (ED-Conv2D) and the cost volume matching with 3D convolution (CVM-Conv3D). Besides, recent studies \cite{Saikia_2019_ICCV, he2019automl} begin to reveal the potential of automated machine learning (AutoML) for neural architecture search (NAS) on stereo matching. In practice, to measure whether a DNN model is applicable in real-world applications, we not only need to evaluate its accuracy on unseen stereo images (whether it can estimate the disparity correctly), but also need to evaluate its time efficiency (whether it can generate the results in real-time). However, existing methods either focus on model accuracy (e.g.,\cite{psmnet2018}\cite{ganet2019}) or on time efficiency (e.g.,\cite{fast2018}\cite{energy_depth_iotj2021}\cite{cyclic2021}), which could make the trained models not applicable to the real-world applications supporting real-time inference on GPU servers or mobile devices with good model accuracy.

 \begin{figure}[t]
    \captionsetup[subfigure]{farskip=1pt}
	\centering
	\subfloat[Left image]
	{
	\adjincludegraphics[width=0.48\linewidth,trim={{.45\width} {.2\width} 0 0},clip]{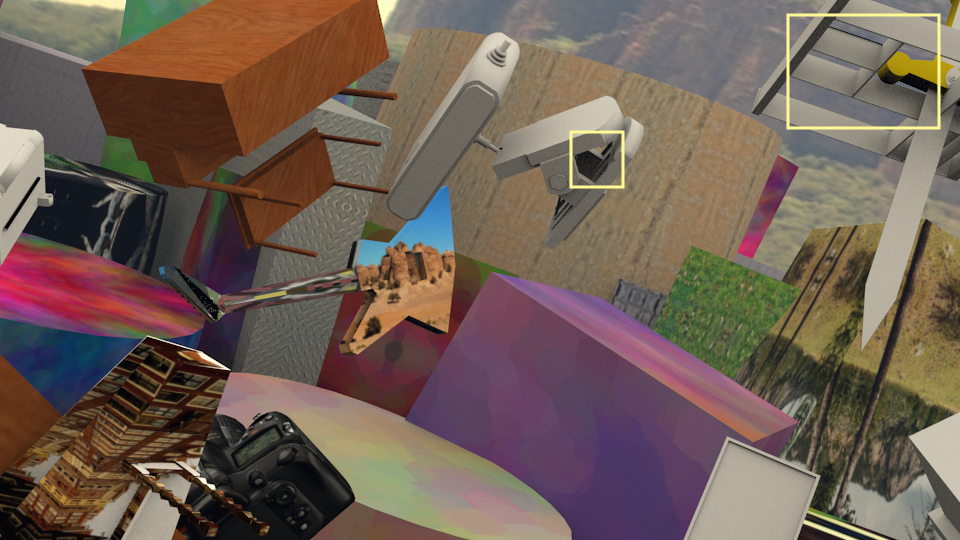}\label{fig:vt_bf_left_rgb}
	}
	\subfloat[Right image]
	{
	\adjincludegraphics[width=0.48\linewidth,trim={{.45\width} {.2\width} 0 0},clip]{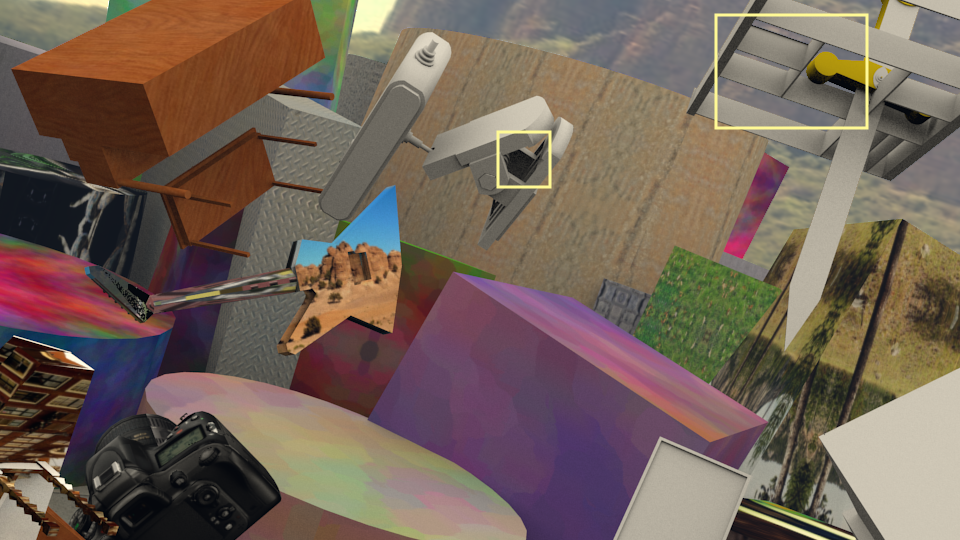}\label{fig:vt_bf_disp_gt}
	}
	\newline
	\subfloat[CRL (0.03 s)\cite{crl2017}]
	{
	\adjincludegraphics[width=0.48\linewidth,trim={{.45\width} {.2\width} 0 0},clip]{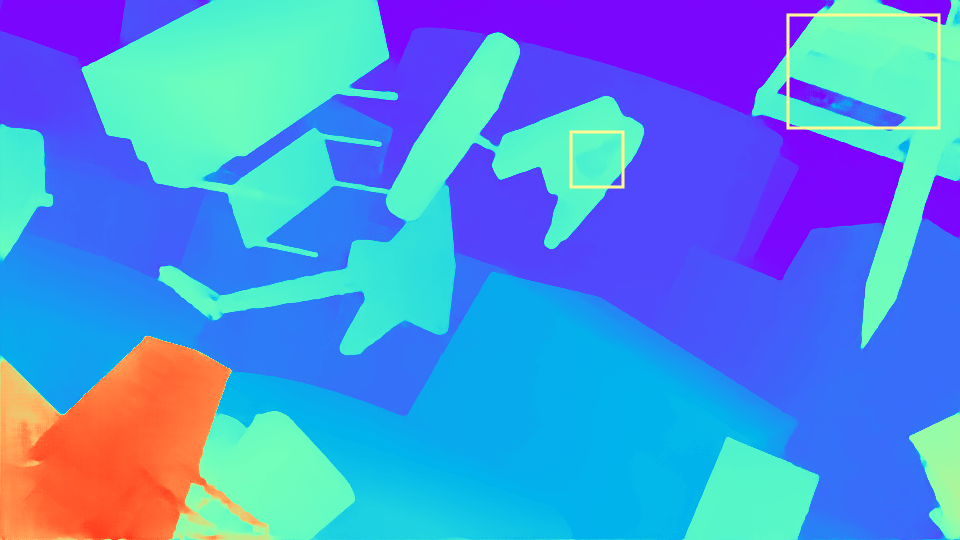}\label{fig:vt_bf_disp_irs}
	}
	\subfloat[GANet (2.29 s) \cite{ganet2019}]
	{
	\adjincludegraphics[width=0.48\linewidth,trim={{.45\width} {.2\width} 0 0},clip]{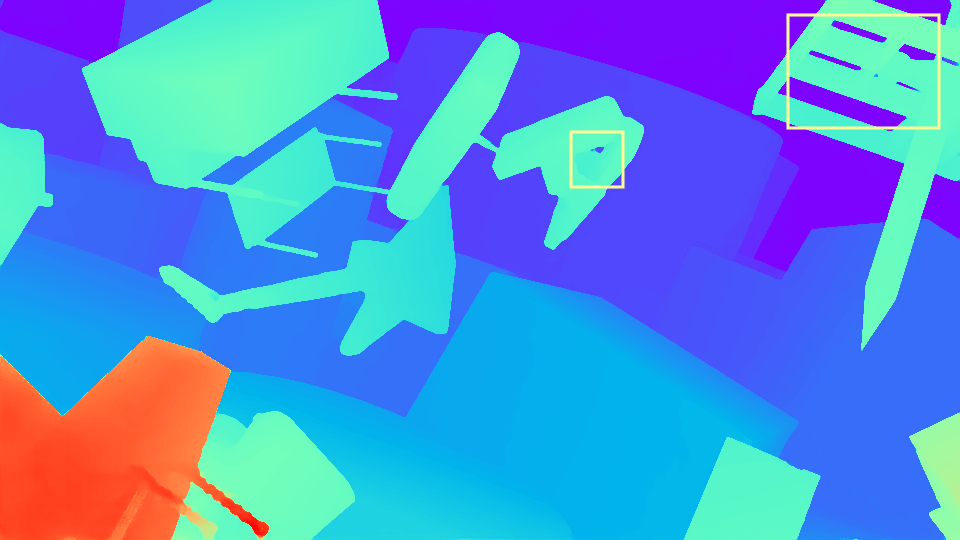}\label{fig:vt_bf_disp_sf}
	}
	\newline
	\subfloat[Our FADNet++ (0.03 s)]
	{
	\adjincludegraphics[width=0.48\linewidth,trim={{.45\width} {.2\width} 0 0},clip]{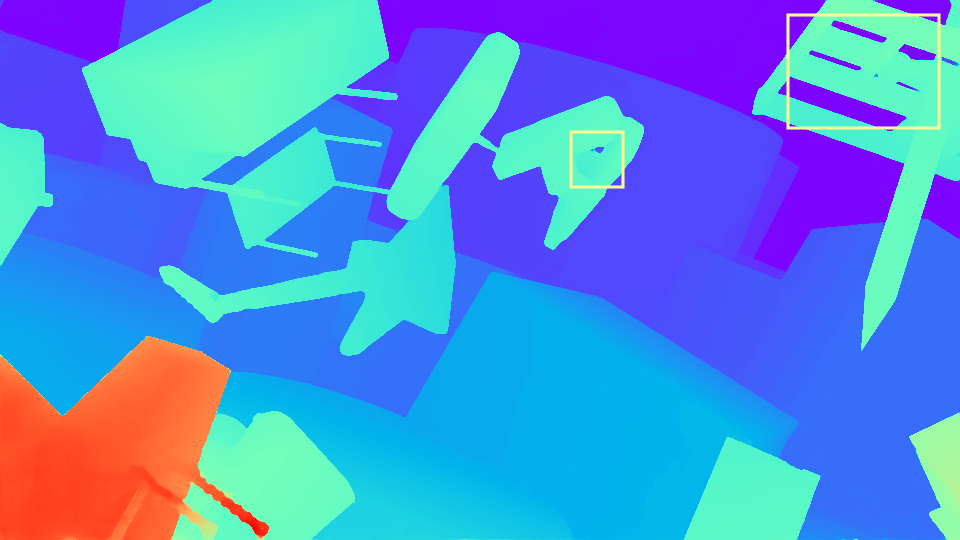}\label{fig:vt_bf_disp_irs}
	}
	\subfloat[Ground truth]
	{
	\adjincludegraphics[width=0.48\linewidth,trim={{.45\width} {.2\width} 0 0},clip]{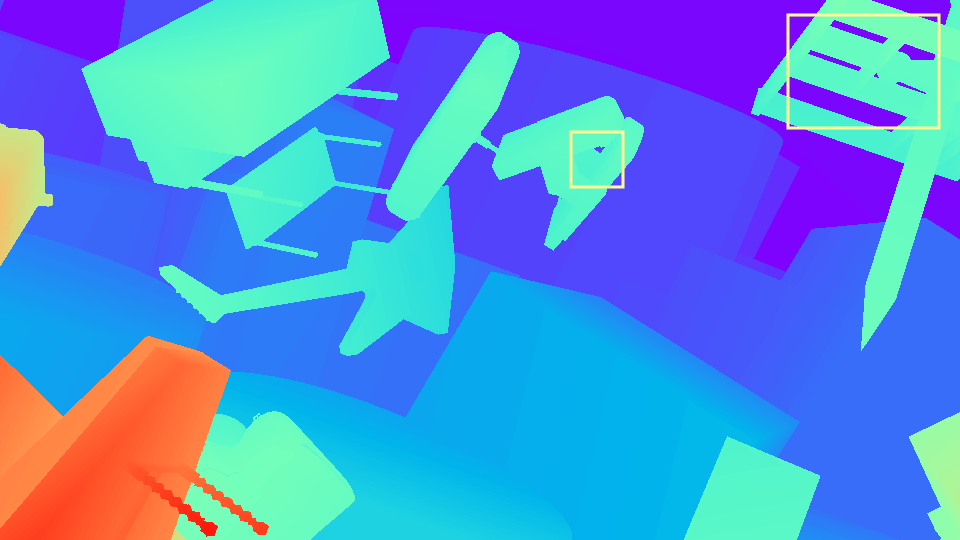}\label{fig:vt_bf_disp_sf}
	}	
	\caption{Performance illustrations of a challenging sample. (a) the left input image. (b) the right input image. (c) result of CRL \cite{crl2017} which runs only 0.03 s but produces wrong disparity values on the shell. (d) result of GANet \cite{ganet2019}, which produces accurate disparity values close to the ground truth but consumes 7.5 GB GPU memory and runs 2.29 s for one stereo image pair. (c) result of our FADNet++, which only consumes 2.3 GB GPU memory and runs 0.03 s to produce the same accurate values. All the data is collected on the Nvidia Tesla V100 GPU.}
	\label{fig:results_preview}
\end{figure} 
\begin{figure}[htbp]
	\centering
	\includegraphics[width=0.98\linewidth]{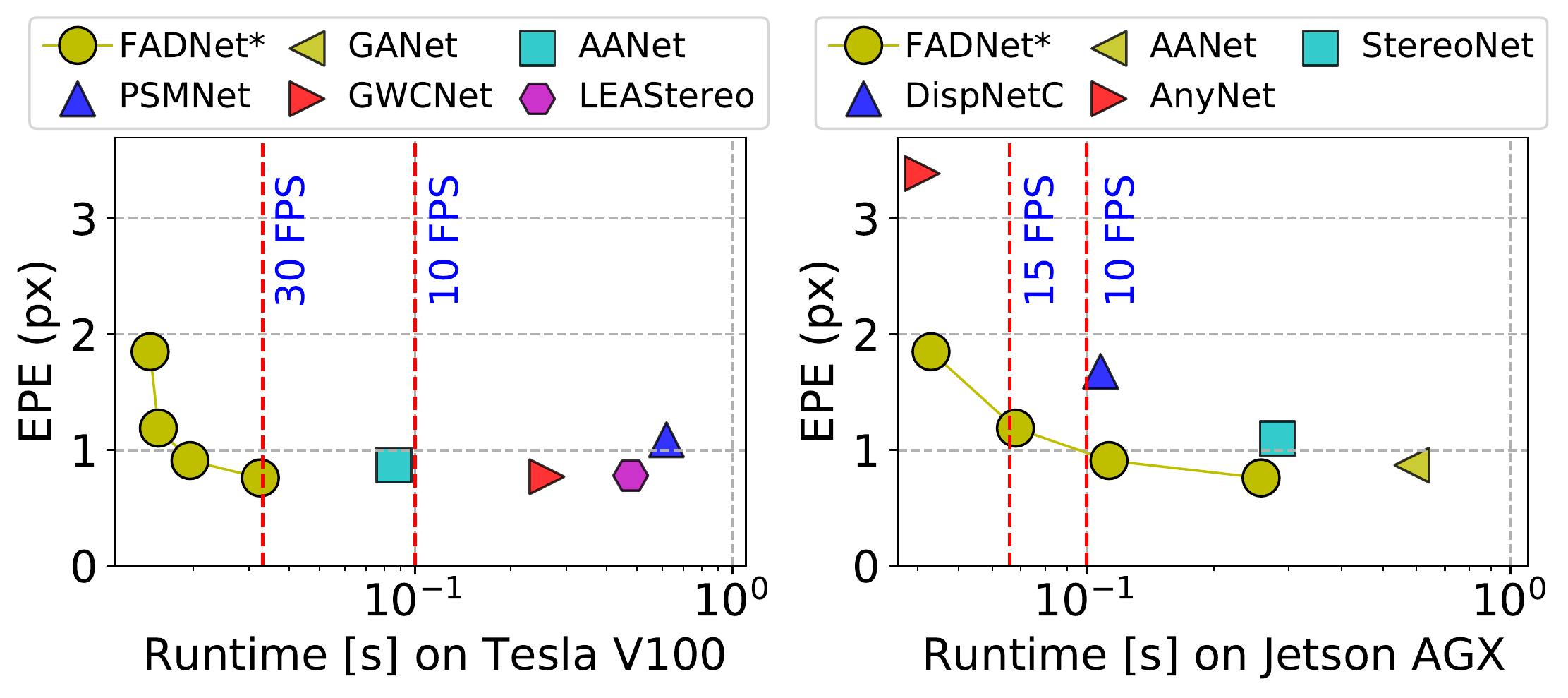}
	\caption{Runtime VS EPE of different methods. On a server GPU Tesla V100, only our FADNet can achieve 30 FPS while the EPE is as low as those CVM-Conv3D models. On a mobile GPU Jetson AGX, our FADNet not only achieves 15 FPS but also produces much lower EPE compared to AnyNet.}
	\label{fig:net_device}
\end{figure}
In ED-Conv2D methods, which are relatively compute-efficient compared to CVM-Conv3D, stereo matching neural networks \cite{zagoruyko2015learning}\cite{zbontar2016stereo}\cite{mayer2016large} are first proposed for end-to-end disparity estimation by exploiting an encoder-decoder structure. The encoder part extracts the features from the input images, and the decoder part predicts the disparity with the generated features. The disparity prediction is optimized as a regression or classification problem using large-scale datasets (e.g., SceneFlow \cite{mayer2016large}) with disparity ground truth. The correlation layer \cite{flownet}\cite{mayer2016large} is then proposed to increase the learning capability of DNNs in disparity estimation, and it has been proved to be successful in learning strong features at multiple levels of scales \cite{flownet}\cite{mayer2016large}\cite{flownet2}\cite{flownet3}\cite{liang2018learning}. To further improve the capability of the models, residual networks \cite{resnet2016}\cite{orhan2017skip}\cite{zhan2019dsnet} are introduced into the architecture of disparity estimation networks since the residual structure enables much deeper network to be easier to train \cite{du2019amnet}. The ED-Conv2D methods have been proven computing efficient, but they cannot achieve very high estimation accuracy~\cite{wang2020fadnet}.

To address the accuracy problem of disparity estimation, researchers have proposed CVM-Conv3D networks to better capture the features of stereo images and thus improve the estimation accuracy \cite{zbontar2016stereo}\cite{kendall2017end}\cite{psmnet2018}\cite{ganet2019}\cite{nie2019multi}. The key idea of the CVM-Conv3D methods is to generate the cost volume by concatenating left feature maps with their corresponding right counterparts across each disparity level \cite{kendall2017end}\cite{psmnet2018}. The features of cost volume are then automatically extracted by 3D convolution layers. 3D operations in DNNs, However, are computing-intensive and hence very slow even with current powerful AI accelerators (e.g., GPUs). Although the 3D convolution based DNNs can achieve state-of-the-art disparity estimation accuracy, they are difficult for deployment due to the very high latency to generate results. On one hand, it requires a large amount of memory to off-load the model, so only a limited set of accelerators (like Nvidia Tesla V100 with 32GB memory) can run these models. On the other hand, it takes several seconds to generate a single result even on a very powerful Nvidia Tesla V100 GPU using CVM-Conv3D models. The memory consumption and the high computation workloads make the CVM-Conv3D methods difficult to be deployed in practice. Therefore, it is crucial to address the accuracy and efficiency problems for real-world applications.

To ease the human efforts of designing an efficient network structure for stereo matching, some recent studies \cite{Saikia_2019_ICCV,cheng2020hierarchical} also take advantages of automated machine learning (AutoML) \cite{he2019automl}, especially the neural architecture search (NAS) technique, to search the optimal set of network operators as well as their connections. However, those state-of-the-art methods are still far from real-time inference even on a server GPU since they are still based on either the complicated network stacking or low-efficient 3D convolution operations. Besides, another series of studies focus on light-weight network structures for fast inference, such as StereoNet \cite{stereonet2018} and AnyNet \cite{anynet2019}. However, the light-weight models significantly sacrifice the model accuracy, especially on some complex realistic datasets, such as KITTI \cite{kitti2015} and Middlebury \cite{scharstein2014high}.

To achieve a practical model in stereo matching, we propose FADNet++ which produces real-time and accurate disparity estimation with configurable networks. 
%FADNet++ can achieve high accuracy while keeping a fast inference speed.
This article is an extension of our previous conference paper \cite{wang2020fadnet}. Similar to the previous FADNet, in FADNet++, we first exploit the multiple stacked 2D-based convolution layers with fast computation, and then we combine state-of-the-art residual architectures to improve the learning capability, and finally we introduce multi-scale outputs for FADNet++ so that it can exploit the multi-scale weight scheduling to improve the training speed. As illustrated in Fig. \ref{fig:results_preview}, our FADNet++ can easily obtain comparable performance as state-of-the-art GANet \cite{ganet2019}, while it runs approximately 70$\times$ faster than GANet and consumes 3$\times$ less GPU memory. Besides, the new FADNet++ advances the previous FADNet in \cite{wang2020fadnet} in three folds. First, we allow configurable variants of FADNet++ to meet different demands of model accuracy and speed. Second, we conduct an extensive comparative study on the model accuracy and speed of different FADNet++ variants during both the training and inference stages. Third, compared to only two stereo datasets and two high-end GPUs in \cite{wang2020fadnet}, we validate our proposed FADNet++ on four stereo datasets and six different GPU platforms from server-level to edge-level. 
%These features enable FADNet++ to efficiently predict the disparity with high accuracy as compared to existing work. 
As shown in Fig. \ref{fig:net_device}, the FADNet++ variants (denoted by ``FADNet*'') can adapt to the platforms of different computing capability. On a server GPU, even the slowest FADNet++ can achieve 30 FPS with a lower EPE than those CVM-Conv3D methods. On a mobile GPU, our FADNet++ can achieve up to 15 frames per second (FPS) with a much lower EPE than the fastest AnyNet \cite{anynet2019}. We make the project of FADNet++ publicly available\footnote{\url{https://github.com/HKBU-HPML/FADNet}}. Our contributions are summarized as follows:
\begin{itemize}
    \item We propose an accurate yet efficient DNN architecture for disparity estimation named FADNet++ (with configurable architecture to support multiple hardware for efficient inference), which achieves comparable prediction accuracy as CVM-Conv3D models and it runs at an order of magnitude faster speed than the 3D-based models.
    \item We develop a multiple rounds training scheme with multi-scale weight scheduling for FADNet++ as well as its variants during training, which improves the training speed yet maintains the model accuracy.
    % \item We conduct quantitative analysis on model parameters and observe that the number of feature maps in FADNet++ can be reduced with little impact on the model accuracy while improving inference speed.
    \item We achieve state-of-the-art accuracy on the Scene Flow dataset with more than 14$\times$ and up to 69$\times$ faster disparity prediction speed than both the NAS-based (LEAStereo \cite{cheng2020hierarchical}) and the human-designed (PSMNet \cite{psmnet2018} and GANet \cite{ganet2019}) models. Besides, by tuning the channel ratios of our FADNet++ to meet the limited computational resources, the variant FADNet-S advances the existing mobile solution, AnyNet \cite{anynet2019}, with much higher prediction accuracy and a competitive inference speed of 15 FPS on the mobile Jetson AGX. 
\end{itemize}

The rest of the paper is organized as follows. We introduce some related work about DNN based solutions to disparity estimation in Section \ref{sec:related_work}. Section \ref{sec:model} introduces the methodology and implementation of our proposed network with configurable size of models. We demonstrate our experimental settings and results in Section \ref{sec:exp}. We finally conclude the paper in Section \ref{sec:conclusion}.
\section{Related Work} \label{sec:related_work}
There exist many studies using deep learning methods in estimating image depth using monocular, stereo and multi-view images. Although monocular vision is low cost and commonly available in practice, it does not explicitly introduce any geometrical constraint, which is important for disparity estimation\cite{Luo_2018_CVPR}. On the contrary, stereo vision leverages the advantages of cross-reference between the left and the right view, and usually shows greater performance and robustness in geometrical tasks. 
Thanks to the rapid and promising development of DNNs, stereo matching also gains considerable credits from DNNs which efficiently extract great feature representation and fit the cost matching function between the left and right view. The early studies mainly focus on optimizing the existing network architectures by enormous hands-on trial-and-error tweaking efforts. 
Besides, recent studies also leverage multi-task learning \cite{segstereo2018,song2020edgestereo,irs2021} to combine other prior vision information and NAS-based methods \cite{cheng2020hierarchical,Saikia_2019_ICCV} to tweak the network structure as well as the operator hyper-parameters (i.e., kernel size and channel number for the convolution layer). 
According to the basic operator (related to the computational efficiency) and the network pipeline, we mainly discuss two branches of network structures for disparity estimation, the ED-Conv2D series and the CVM-Conv3D series.

\subsection{Disparity Estimation with ED-Conv2D CNNs}
In the ED-Conv2D series, end-to-end architectures with mainly convolution layers \cite{mayer2016large}\cite{crl2017} are proposed for disparity estimation, which use two stereo images as input and generate the disparity directly and the disparity is optimized as a regression task. This is achieved by adopting large U-shape encoder-decoder networks with 2D convolutions to predict the disparity map. However, the models that are pure 2D CNN architectures are difficult to capture the matching features such that the estimation results are not good. To address the problem, the correlation layer which can express the relationship between left and right images is introduced in the end-to-end architecture (e.g., DispNetCorr1D \cite{mayer2016large}, FlowNet \cite{flownet}, FlowNet2 \cite{flownet2}, DenseMapNet \cite{flownet3}). The correlation layer significantly increases the estimating performance compared to the pure CNNs, but existing architectures are still not accurate enough for production. Furthermore, CRL \cite{crl2017} and FADNet \cite{wang2020fadnet} introduce the idea of residual learning \cite{resnet2016} to conduct efficient disparity refinement in a coarse-to-fine manner. Liang et al. \cite{iresnet2021} apply the similar idea of them but with constructing multi-scale cost volumes from the feature pyramid. Although those existing ED-Conv2D methods enjoy the high model inference efficiency, they usually fail to produce satisfactory results in some challenging scenarios. Besides, some studies leverage multi-task learning to incorporates other visual information, such as edge cues \cite{song2020edgestereo} and semantic segmentation \cite{segstereo2018}, to promote the accuracy of the textureless regions, detailed structures and small objects. 

\subsection{Disparity Estimation with CVM-Conv3D CNNs}
The CVM-Conv3D CNNs are further proposed to increase the estimation performance \cite{zbontar2016stereo}\cite{kendall2017end}\cite{psmnet2018}\cite{ganet2019}\cite{nie2019multi}, which leverage the concept of semi-global matching \cite{hirschmuller2007stereo} to learn disparities from a 4D cost volume. The cost volume is mainly constructed by concatenating left feature maps with their corresponding right counterparts across each disparity level \cite{kendall2017end}\cite{psmnet2018}, and the features of the generated cost volumes can be learned by 3D convolution layers. The CVM-Conv3D CNNs can automatically learn to regularize the cost volume, which have achieved state-of-the-art accuracy of various datasets. However, the key limitation of the 3D based CNNs is their extremely high computation resource requirements. For example, training GANet \cite{ganet2019} with the Scene Flow \cite{mayer2016large} dataset takes weeks even using very powerful Nvidia Tesla V100 GPUs. Even they achieve good accuracy, it is difficult to deploy due to their very low time efficiency. 
Thus, recent research proposes some optimization solutions, such as cost volume compression by grouping \cite{gwcnet2019}, efficient search space pruning \cite{deeppruner2019} and corporative learning of multi-scale features \cite{xu2020aanet}. However, the fastest AANet \cite{xu2020aanet} among all CVM-Conv3D CNNs only runs 12 FPS even on a great Tesla V100 GPU and is still far from real-time inference on other low-end devices. Besides, to lessen the effort dedicated to designing network architectures, automated machine learning (AutoML) \cite{he2019automl} especially neural architecture search (NAS) \cite{nas2019,liu2018darts,cai2020once}, has also been applied to stereo matching in \cite{cheng2020hierarchical,Saikia_2019_ICCV,searchdense2018} and successfully achieved the leader accuracy and generalization in several benchmarks. However, the low time efficiency and high memory footprint of those 3D-conv based architectures still remain. To this end, we propose a fast and accurate DNN model for disparity estimation.

\section{Approach} \label{sec:model}
\begin{figure*}[htbp]
	\centering
	\includegraphics[width=0.96\linewidth]{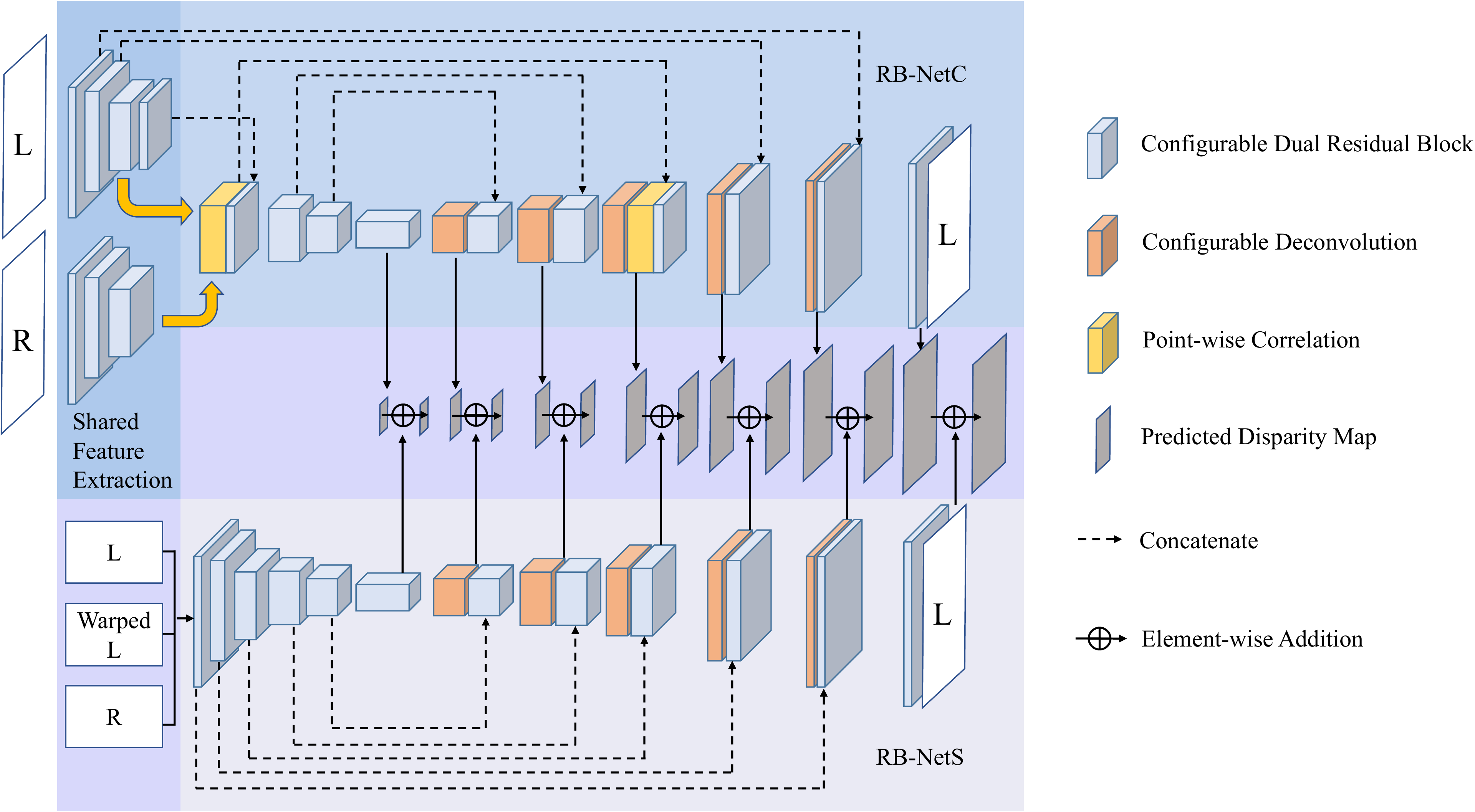}
	\caption{The model structure of our proposed FADNet++. ``Configurable'' indicates that the channel numbers of the convolution/deconvolution layers can be modified by a tunable ratio (discussed in Section \ref{subsec:configurable}) to control the overall model size. ``L'' indicates the left input image, and ``R'' indicates the right input image. ``Warped L'' indicated the aligned left image produced by warping the right image with the initial predicted disparity map of RB-NetC. The sizes of different predicted disparity maps reflect their scales in the network.}
	\label{fig:fadnet}
\end{figure*}
\subsection{Model Design and Implementation}
Our proposed FADNet++ exploits the structure of DispNetC \cite{mayer2016large} as a backbone, but it is extensively reformed to take care of both accuracy and inference speed, which is lacking in existing studies. We introduce four novel components in FADNet++ to enable its good generalization ability and fast inference speed with configurable size for different hardware. 1) We first change the structure in terms of branch depth and layer type by introducing two new modules, residual block and point-wise correlation; 2) Then we exploit the multi-scale residual learning strategy for training the refinement network; 3) We design the model to be configurable (with a scaling ratio) to balance the accuracy and inference speed. 4) Finally, a loss weight training schedule is used to train the network in a coarse-to-fine manner. 

\subsection{Residual Block and Point-wise Correlation}\label{subsec:pwc}
\begin{figure}[htbp]
	\centering
	\includegraphics[width=0.96\linewidth]{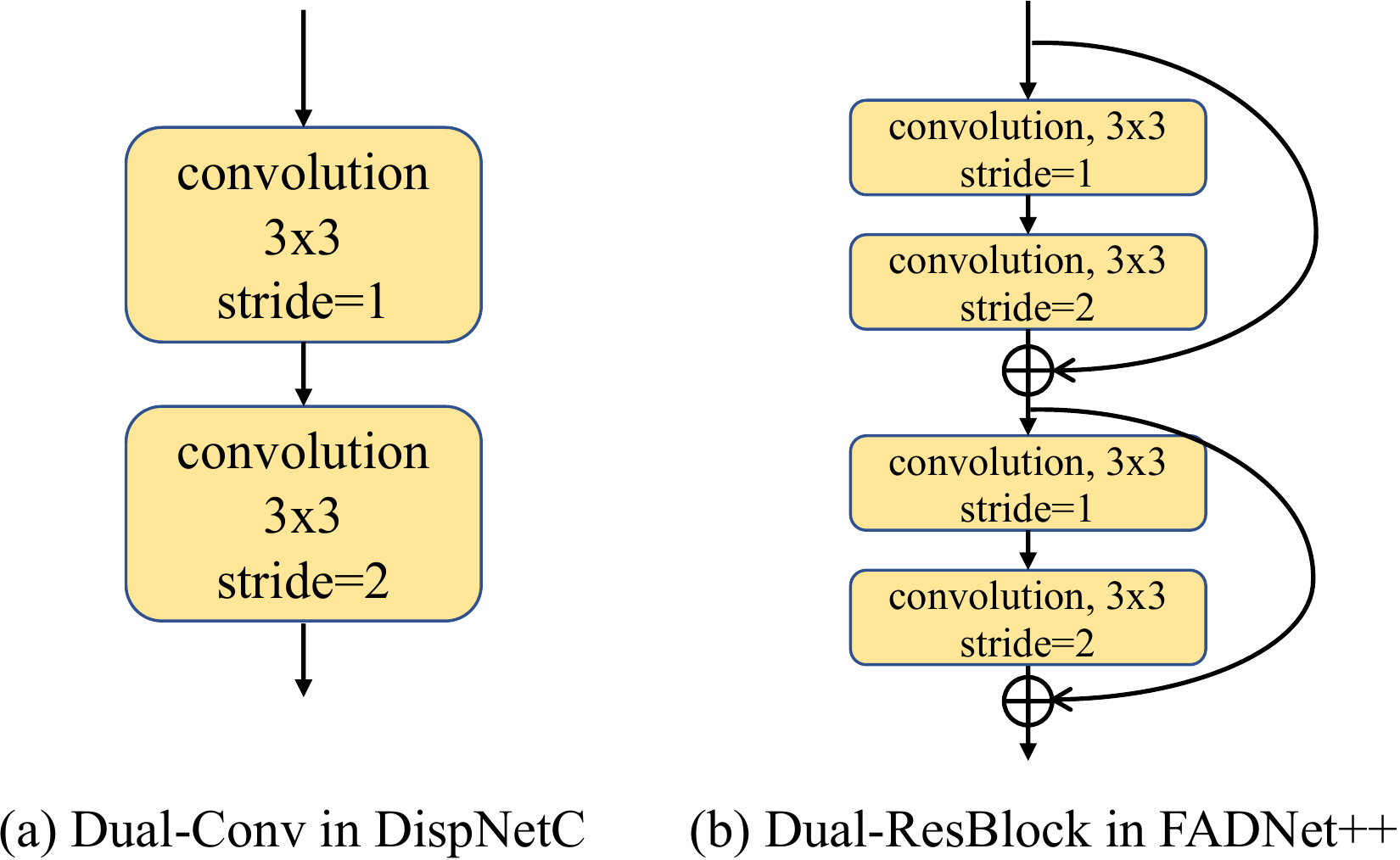}
	\caption{the original two-layer convolutions (Dual-Conv) in DispNetC \cite{mayer2016large}, while the right part shows the Dual-ResBlock module applied in our FADNet++.}
	\label{fig:resblock}
\end{figure}

DispNetC and DispNetS which are both from the study in \cite{mayer2016large} basically use an encoder-decoder structure equipped with five feature extraction and down-sampling layers and five feature deconvolution layers. While conducting feature extraction and down-sampling, DispNetC and DispNetS first adopt a convolution layer with a stride of 1 and then a convolution layer with a stride of 2 so that they consistently shrink the feature map size by half. We call the two-layer convolutions with size reduction as Dual-Conv, as shown in Fig. \ref{fig:resblock}(a). DispNetC equipped with Dual-Conv modules and a correlation layer finally achieves an end-points error (EPE) of 1.68 on the SceneFlow dataset~\cite{mayer2016large}, as reported in \cite{mayer2016large}.

The residual block originally derived in \cite{resnet2016} for image classification tasks is widely used to learn robust features and train a very deep network. The residual block can well address the gradient vanish problem when training very deep networks. Thus, we replace the convolution layer in the Dual-Conv module by the residual block to construct a new module called Dual-ResBlock, as shown in Fig. \ref{fig:resblock}(b). With Dual-ResBlock, we can make the network deeper without training difficulty as the residual block allows us to train very deep models. Therefore, we further increase the number of feature extraction and down-sampling layers from five to seven. Finally, DispNetC and DispNetS are evolving to two new networks with better learning ability, which are called RB-NetC and RB-NetS respectively, as shown in Fig. \ref{fig:fadnet}. 

One of the most important contributions of DispNetC is the correlation layer, which targets at finding correspondences between the left and right images. Given two multi-channel
feature maps $\textbf{f}_1,\textbf{f}_2$ with $w, h$ and $c$ as their width, height and number of channels, the correlation layer calculates the cost volume of them using Eq. \eqref{eq:corr}.
\begin{align}
    c(\textbf{x}_1,\textbf{x}_2)=\sum_{\textbf{o} \in [-k, k]\times [-k,k]}\langle\textbf{f}_1(\textbf{x}_1 + \textbf{o}), \textbf{f}_2(\textbf{x}_2 + \textbf{o}) \rangle, \label{eq:corr}
\end{align}
where $k$ is the kernel size of cost matching, $\textbf{x}_1$ and $\textbf{x}_2$ are the centers of two patches from $\textbf{f}_1$ and $\textbf{f}_2$ respectively. Computing all patch combinations involves $c\times K^2 \times w^2 \times h^2$ multiplication and produces a cost matching map of $w \times h$. Given a maximum searching range $D$, we fix $\textbf{x}_1$ and shift the $\textbf{x}_2$ on the x-axis direction from $-D$ to $D$ with a stride of two. Thus, the final output cost volume size becomes $w\times h \times D$. 

However, the correlation operation assumes that each pixel in the patch contributes equally to the point-wise convolution results, which may lost the ability to learn more complicated matching patterns. Here we propose point-wise correlation composed of two modules. The first module is a classical convolution layer with a kernel size of $3\times3$ and a stride of $1$. The second one is an element-wise multiplication which is defined by Eq. \eqref{eq:pw_corr}.
\begin{align}
    c(\textbf{x}_1,\textbf{x}_2)=\sum{\langle\textbf{f}_1(\textbf{x}_1), \textbf{f}_2(\textbf{x}_2) \rangle}, \label{eq:pw_corr}
\end{align}
where we remove the patch convolution manner from Eq. \eqref{eq:corr}. Note that the maximum search range for the original image resolution should not be larger than the maximum valid disparity. For example, in the SceneFlow dataset, its maximum valid disparity is 192, and the correlation layer of our FADNet++ is put after the third Dual-ResBlock, of which the output feature resolution is 1/8. So a proper searching range value should not be less than 192/8=16. We set a marginally larger value 20. We also test some other values, such as 10 and 40, which do not surpass the version of using 20. The reason is that applying too small or large search range value may lead to under-fitting or over-fitting. 

Table \ref{tab:res_corr} lists the accuracy improvement brought by applying the proposed Dual-ResBlock and point-wise correlation. To simplify the validation experiment, we train them using the same SceneFlow \cite{mayer2016large} dataset for only 20 epochs, which is different from the complete training scheme in Section \ref{sec:exp}. It is observed that RB-NetC outperforms DispNetC with a much lower EPE, which indicates the effectiveness of the residual structure. We also notice that setting a proper searching range value of the correlation layer helps further improve the model accuracy. 
\begin{table}[!ht]
	\centering
	\caption{Model accuracy improvement of Dual-ResBlock and point-wise correlation with different $D$.}
	\label{tab:res_corr}
% 	\scriptsize{
		\begin{tabular}{|c|c|c|c|} \hline
			\textbf{Model} & $D$ & Training EPE & Test EPE \\ \hline\hline
			DispNetC & 20 & 2.89 & 2.80 \\ \hline
			RB-NetC &10 & 2.28 & 2.06 \\ \hline
			RB-NetC &20 & 2.09 & 1.76 \\ \hline
			RB-NetC &40 & 2.12 & 1.83 \\ \hline
		\end{tabular}
% 	}	
\end{table}

\subsection{Multi-Scale Residual Learning}
Instead of directly stacking DispNetC and DispNetS sub-networks to conduct disparity refinement procedure \cite{flownet3}, we apply the multi-scale residual learning firstly proposed by \cite{crl2017}. The basic idea is that the second refinement network learns the disparity residuals and accumulates them into the initial results generated by the first network, instead of directly predicting the whole disparity map. In this way, the second network only needs to focus on learning the highly nonlinear residual, which is effective to avoid gradient vanishing. Our final FADNet++ is formed by stacking RB-NetC and RB-NetS with multi-scale residual learning, which is shown in Fig. \ref{fig:fadnet}. 

As illustrated in Fig. \ref{fig:fadnet}, the upper RB-NetC takes the left and right images as input and produces disparity maps at a total of 7 scales, denoted by $c_{s}$, where $s$ is from 0 to 6. The bottom RB-NetS exploits the inputs of the left image, right image, and the warped left images to predict the residuals. The generated residuals (denoted by $r_{s}$) from RB-NetS are then accumulated to the prediction results by RB-NetC to generate the final disparity maps with multiple scales ($s=0,1,...,6$). Thus, the final disparity maps predicted by FADNet++, denoted by $\hat{d_{s}}$, can be calculated by
\begin{align}
    \hat{d_{s}} = c_{s} + r_{s}, 0 \leq s \leq 6. \label{eq:d_r_sum}
\end{align}

\subsection{Configurable Network Size} \label{subsec:configurable}
Although the recent state-of-the-art models, such as PSMNet \cite{psmnet2018}, GANet \cite{ganet2019}, LEAStereo \cite{cheng2020hierarchical} and our previous FADNet \cite{wang2020fadnet}, produce decent accuracy of disparity estimation, the practicability on computing devices of different computational capability, especially those low-end mobile ones, has not yet been extensively studied. Recently, AnyNet \cite{anynet2019} reduced the inference overhead of stereo matching by alternatively refining the disparity map in a coarse-to-fine manner according to the target device, and made it possible to be deployed on a mobile Jetson TX2 platform with over 20 FPS. However, the low-level features, which are important to recover the object details and boundaries, could be discarded to keep a high inference speed on a low-end device. Prior to AnyNet, we keep all the features from low to high scales but make the channel numbers of convolution/deconvolution layers configurable so that we can balance the model accuracy and inference speed. Our design has three advantages. First, the network size can be easily controlled by two ratio parameters, which is proved to be simple yet effective in our experiments. Second, the variants of different configurations still share the overall network structure of FADNet++ instead of dropping some layers/modules (as adopted in \cite{stereonet2018}) or some scales (as adopted in \cite{anynet2019}) such that the benefits of the FADNet++ backbone can be maintained. Third, the configurable ratio is convenient in terms of balancing the accuracy and performance under different application requirements. 

In our proposed FADNet++, RB-NetC and RB-NetS have the same number of layers in their decoder and encoder parts, respectively. Assume that the encoder part has $E$ layers and the decoder part has $D$ layers. The $i^{\text{th}}$ layer in the encoder is denoted by $l_{i}^{E}$. The $i^{\text{th}}$ layer in the decoder is denoted by $l_{i}^{D}$. For each convolution layer, we have a basic channel number denoted by $\hat{C}$, which also indicates the minimum channels. Then we introduce two ratios, E-Ratio for encoders and D-Ratio for decoders, to conveniently configure the model size. By assigning different values for E-Ratio and D-Ratio, we are able to construct a set of FADNet++ variants. We list some of them in Table \ref{tab:fadnet_variants}. The channel number of each convolution layer can be calculated by
\begin{subequations}
\begin{align}
	C_{l_{i}^{E}} &= \hat{C}_{l_{i}^{E}} \times \text{E-Ratio} \\
	C_{l_{i}^{D}} &= \hat{C}_{l_{i}^{D}} \times \text{D-Ratio}
\end{align}
\end{subequations}
The feature of configurable network size obviously promotes the flexibility of FADNet++ in terms of network parameters as well as the model inference speed. We will further evaluate its effectiveness and efficiency in Section \ref{sec:exp} by deploying different variants to a wide range of computing devices. On the one hand, on a server GPU, the full FADNet++ outperforms those expensive CVM-Conv3D methods with slightly better accuracy and a considerable margin of model speed. On the other hand, on a mobile device, the shrinking FADNet-T beats the real-time AnyNet with equivalent model speed but much lower prediction errors. 

\begin{table}[ht]
	\centering
	\caption{FADNet++ variants of different configurations. }
	\label{tab:fadnet_variants}
	\small{
	\begin{tabular}{|c|c|c|c|} \hline
		Network & E-Ratio & D-Ratio & Params [M] \\ \hline\hline
		FADNet++ & 16 & 16 & 124.38 \\ 
		FADNet-M & 8 & 8 & 31.15 \\ 
		FADNet-S & 4 & 4 & 7.82 \\
		FADNet-T & 2 & 1 & 1.65\\ \hline
	\end{tabular}
	}
\end{table}

\subsection{Loss Function Design}\label{subsec:loss}
Given a pair of stereo RGB images, our FADNet++ takes them as input and produces seven disparity maps at different scales. Assume that the input image size is $H \times W$. The dimension of the seven scales of the output disparity maps are $H \times W$, $\frac{1}{2}H \times \frac{1}{2}W$, $\frac{1}{4}H \times \frac{1}{4}W$, $\frac{1}{8}H \times \frac{1}{8}W$, $\frac{1}{16}H \times \frac{1}{16}W$, $\frac{1}{32}H \times \frac{1}{32}W$, and $\frac{1}{64}H \times \frac{1}{64}W$ respectively. To train FADNet++ in an end-to-end manner, we adopt the pixel-wise smooth L1 loss between the predicted disparity map and the ground truth using  
\begin{align}
    L_s(d_s, \hat{d_s})=\frac{1}{N}\sum_{i=1}^{N}{smooth}_{L_1}(d_{s}^i - \hat{d_{s}^i}), \label{eq:smooth_l1}
\end{align}
where $N$ is the number of pixels of the disparity map, $d_s^i$ is the $i^th$ element of $d_s\in \mathcal{R}^N$ and 
\begin{align}
    {smooth}_{L_1}(x)=
    \begin{cases}
    0.5x^2,& \text{if } |x| < 1\\
    |x|-0.5,              & \text{otherwise}.
\end{cases}
\end{align}
Note that $d_s$ is the ground truth disparity of scale $\frac{1}{2^s}$ and $\hat{d_s}$ is the predicted disparity of scale $\frac{1}{2^s}$. The loss function is separately applied in the seven scales of outputs, which generates seven loss values. The loss values are then accumulated with loss weights. 

\begin{table}[!ht]
	\centering
	\caption{Multi-scale loss weight scheduling.}
	\label{tab:loss_weights}
% 	\scriptsize{
		\begin{tabular}{|c|c|c|c|c|c|c|c|} \hline
		\textbf{Round}	& $w_0$ & $w_1$ & $w_2$ & $w_3$ & $w_4$ & $w_5$ & $w_6$ \\ \hline\hline
			1 & 0.32 & 0.16 & 0.08 & 0.04 & 0.02 & 0.01 & 0.005 \\ \hline
			2 & 0.6 & 0.32 & 0.08 & 0.04 & 0.02 & 0.01 & 0.005 \\ \hline
			3 & 0.8 & 0.16 & 0.04 & 0.02 & 0.01 & 0.005 & 0.0025 \\ \hline
			4 & 1.0 & 0 & 0 & 0 & 0 & 0 & 0\\ \hline
		\end{tabular}
% 	}	
\end{table}

The loss weight scheduling technique which is initially proposed in \cite{mayer2016large} is useful to learn the disparity in a coarse-to-fine manner. Instead of just switching on/off the losses of different scales, we apply different non-zero weight groups for tackling different scale of disparity. Let $w_s$ denote the weight for the loss of the scale of $s$. The final loss function is
\begin{equation}
    L=\sum_{s=0}^{6}w_sL_s(d_s,\hat{d_s}).
\end{equation}
The specific setting is listed in Table \ref{tab:loss_weights}. Totally there are seven scales of predicted disparity maps. At the beginning, we assign low-value weights for those large scale disparity maps to learn the coarse features. Then we increase the loss weights of large scales to let the network gradually learn the finer features. Finally, we deactivate all the losses except the final predict one of the original input size. With different rounds of weight scheduling, the evaluation EPE is gradually increased to the final accurate performance which is shown in Table \ref{tab:weightsresults} on the SceneFlow dataset.

\begin{table}[!ht]
	\centering
	\caption{Model accuracy with different rounds of weight scheduling.}
	\label{tab:weightsresults}
		\addtolength{\tabcolsep}{-2.0pt}
		\begin{tabular}{|c|c|c|c|c|c|} \hline
		    Network & Round & \# Epochs & \makecell{Training \\ EPE} & \makecell{Test \\ EPE} & Improvement (\%) \\ \hline\hline
			\multirow{4}{*}{FADNet++} & 1 & 20 & 1.45 & 1.28 & - \\ 
			& 2 & 20 & 1.07 & 0.96 & 25.0 \\ 
			& 3 & 20 & 0.91 & 0.89 & 7.3 \\ 
			& 4 & 30 & 0.74 & 0.76 & 14.6 \\ \hline
			\multirow{4}{*}{FADNet-M} & 1 & 20 & 1.61 & 1.38 & - \\ 
			& 2 & 20 & 1.31 & 1.19 & 13.8 \\ 
			& 3 & 20 & 1.16 & 1.02 & 14.3 \\ 
			& 4 & 30 & 0.97 & 0.91 & 10.8 \\ \hline
			\multirow{4}{*}{FADNet-S} & 1 & 20 & 2.10 & 1.91 & - \\ 
			& 2 & 20 & 1.72 & 1.54 & 19.4 \\ 
			& 3 & 20 & 1.58 & 1.35 & 12.3 \\ 
			& 4 & 30 & 1.47 & 1.19 & 11.9 \\ \hline
			\multirow{4}{*}{FADNet-T} & 1 & 20 & 3.10 & 2.52 & - \\ 
			& 2 & 20 & 2.65 & 2.16 & 14.3 \\ 
			& 3 & 20 & 2.49 & 2.11 & 2.3 \\ 
			& 4 & 30 & 2.25 & 1.83 & 13.3 \\ \hline
		\end{tabular}
		\begin{tablenotes}
    	    \item Note: ``Improvement'' indicates the test EPE decrease of the current round of weight schedule over its previous.
        \end{tablenotes}
\end{table}
Table \ref{tab:weightsresults} lists the model accuracy improvements (average 13.3\% and up to 25.0\% among all the rounds) brought by the multiple round training of four loss weight groups. For each tested network, it is observed that both the training and testing EPEs are decreased smoothly and close, which indicates good generalization and advantages of our training strategy.
\section{Experimental Studies} \label{sec:exp}
In this section, we present the experimental studies to show the effectiveness of our FADNet++. We first demonstrate the accuracy of our proposed networks on different datasets compared to existing state-of-the-art methods. Then we present the inference performance on some popular inference GPUs (including server GPUs and mobile GPUs) to show that our networks are able to support real-time disparity estimation (i.e., not less than 30FPS).

\subsection{Experimental Setup}
\textbf{Testbed}. For model training, we use four Nvidia Tesla V100-PCIe GPUs to train all compared models. For model inference, to cover various types of inference GPUs, we choose a desktop-level Nvidia RTX2070 GPU and two server-level Nvidia GPUs (i.e., Tesla P40 and Tesla T4) to measure the inference speed. We also choose two mobile GPUs including Jetson TX2 and Jeston AGX to evaluate the inference speed. The details of the training and inference servers are shown in Table~\ref{table:testbed-servers}, and the inference mobile devices are shown in Table~\ref{table:testbed-devices}. In terms of software that are related to the time performance, the server is installed with GPU Driver-440.36, CUDA-10.2, and PyTorch-1.4.0 with cuDNN-7.6.

\begin{table}[!ht]
	\centering
		\caption{The inference server configuration.}
		\label{table:testbed-servers}
		\begin{tabular}{|c|c|c|c|c|}
			\hline
			 & \multirow{2}{*}{Training Server} & \multicolumn{3}{c|}{Inference Servers} \\\cline{3-5}
			 & & IS1 & IS2 & IS3 \\\hline\hline
			GPU & Tesla V100$\times$ 4 & RTX2070 & Tesla P40 & Tesla T4 \\\hline
% 			CPU & Dual Intel(R) Xeon(R) Gold 6230 CPU @ 2.10GHz \\
			Memory & 512GB & 32GB & 256GB & 64GB \\\hline
			OS & CentOS7.2 & \multicolumn{3}{c|}{Ubuntu16.04}\\\hline
		\end{tabular}
\end{table}

\begin{table}[!ht]
	\centering
		\caption{The mobile platform configuration.}
		\label{table:testbed-devices}
		\begin{tabular}{|c|c|c|}
			\hline
			& Jetson TX2 & Jetson AGX \\ \hline\hline
			CPU & \makecell{2-Core NVIDIA Denver \\ +4-Core ARM Cortex-A57}  & 8-Core ARM v8.2 \\ \hline
			GPU & 256-Core Pascal & 512-Core Volta \\ \hline
			Memory & 8GB & 32GB \\\hline
			OS & \multicolumn{2}{c|}{Ubuntu 18.04.5, JetPack 4.4} \\\hline
		\end{tabular}
\end{table}

\begin{figure*}[htbp]
	\centering
	\subfloat[SceneFlow]
	{
	\includegraphics[width=0.24\linewidth]{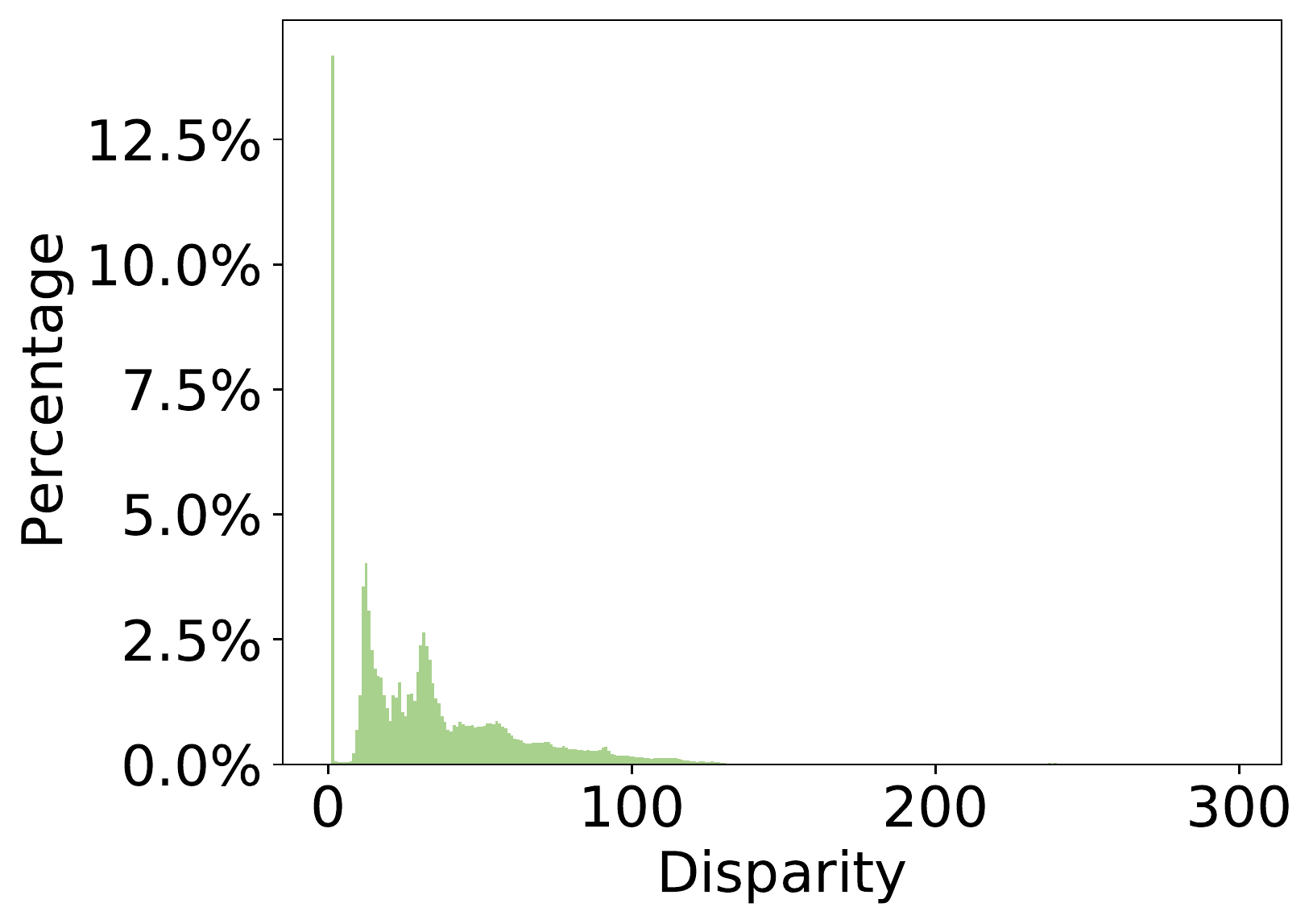}}
	\subfloat[KITTI2015]
	{
	\includegraphics[width=0.24\linewidth]{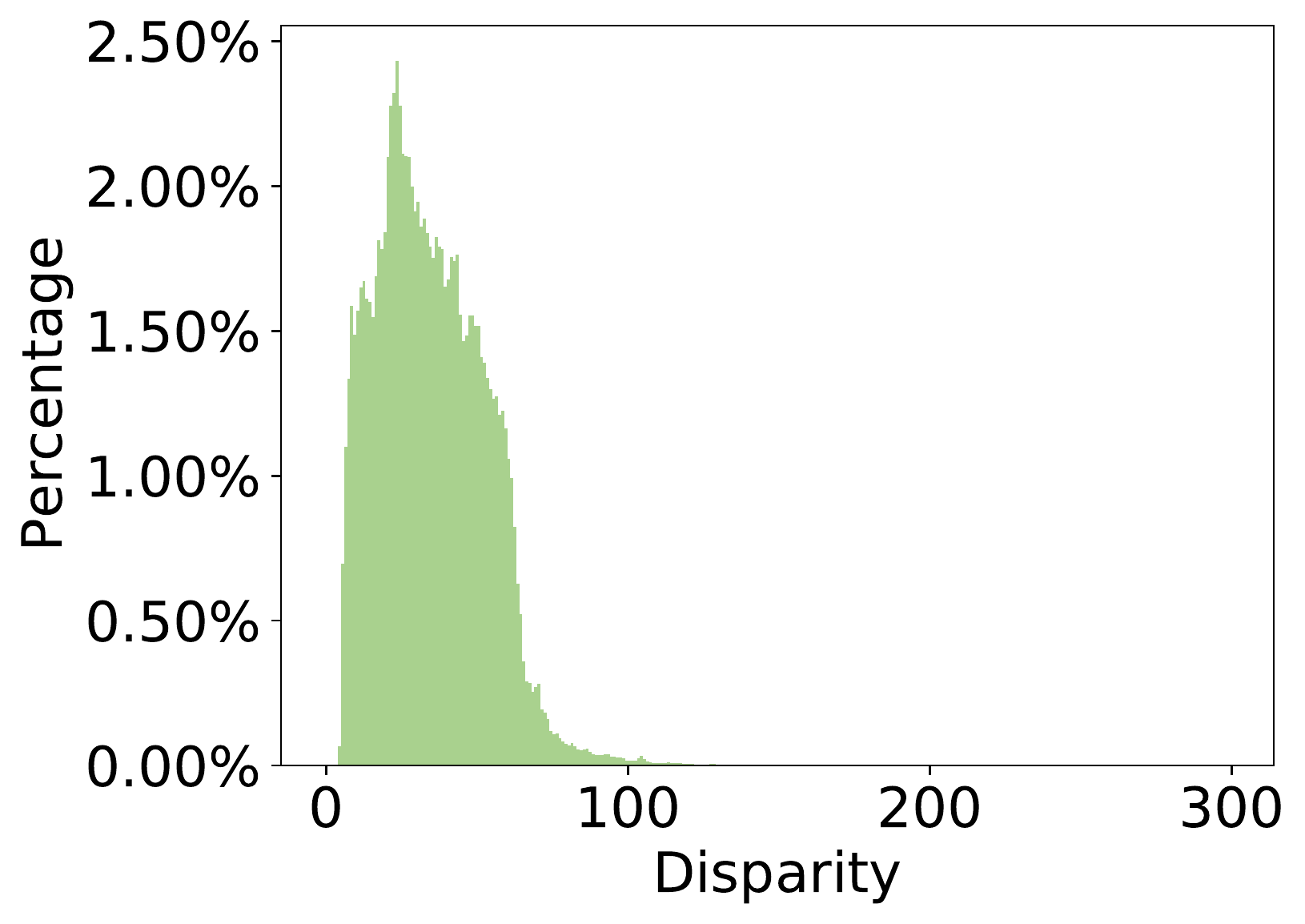}}
	\subfloat[Middlebury]
	{
	\includegraphics[width=0.24\linewidth]{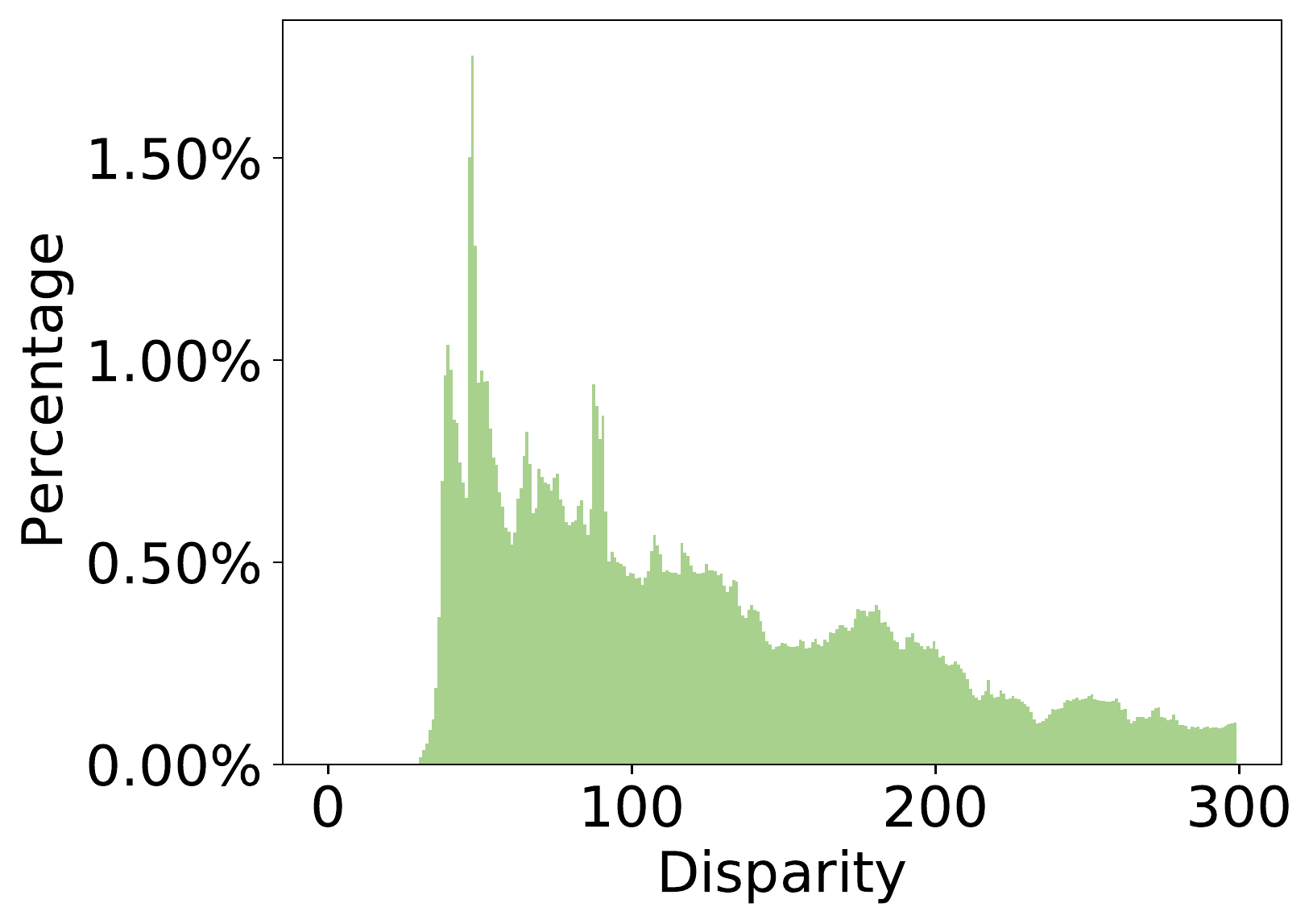}}
	\subfloat[ETH3D]
	{
	\includegraphics[width=0.24\linewidth]{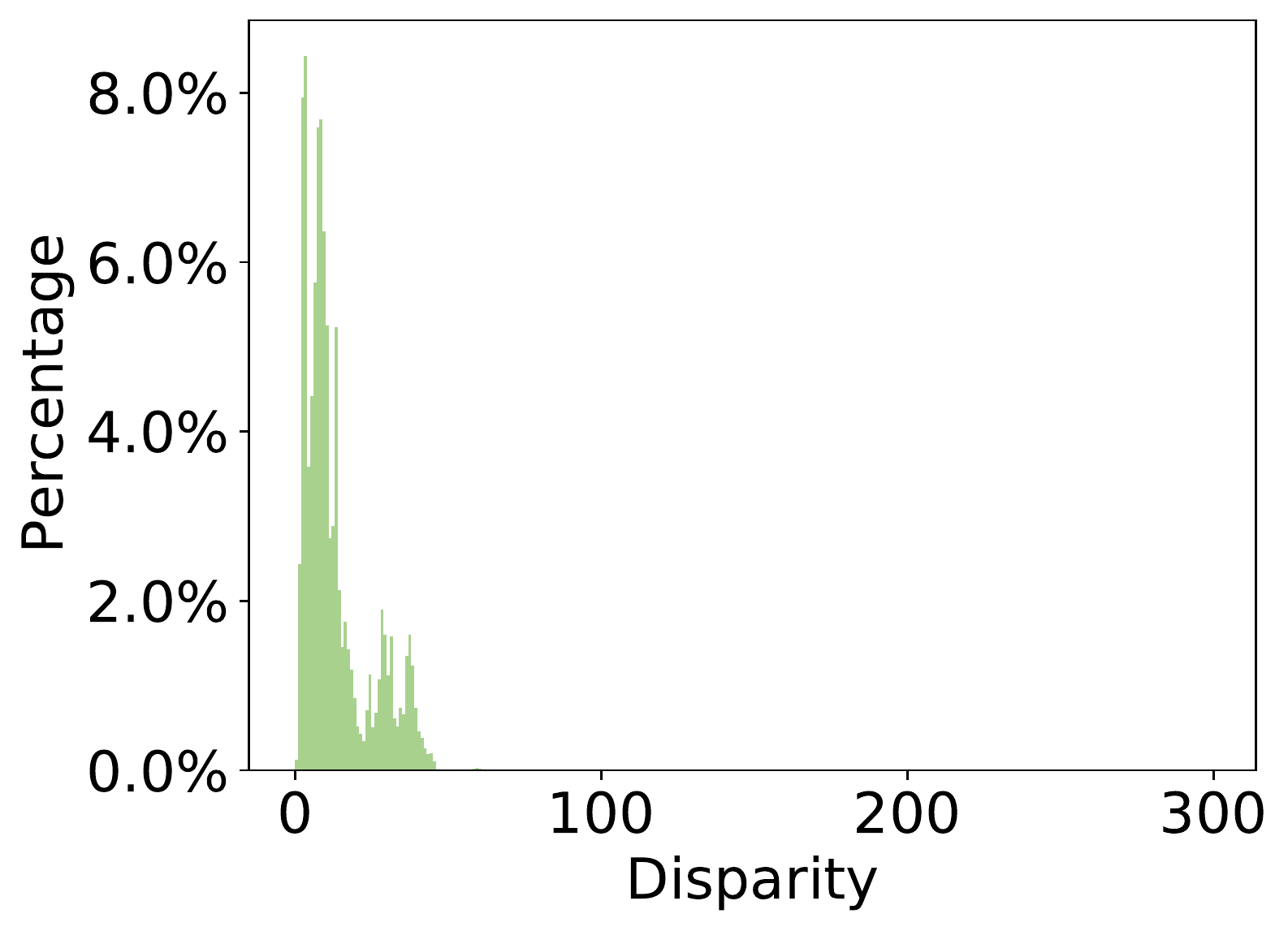}}
	\caption{Disparity distribution in different datasets. Note that zero disparities are excluded.}
	\label{fig:distribution}
\end{figure*}

\textbf{Datasets}. To cover a range of scenarios in disparity estimation, we use many popular public datasets, including Middlebury 2014 (M2014) \cite{scharstein2014high}, KITTI 2015 (K2015) \cite{kitti2015}, ETH3D 2017 (ETH3D) \cite{eth3d2017}, and SceneFlow (SF) \cite{mayer2016large}, to evaluate the performance of different algorithms. The details of the datasets are shown in Table~\ref{table:datasets}.
\begin{table}[!ht]
	\centering
		\caption{The evaluated datasets.}
		\label{table:datasets}
		\addtolength{\tabcolsep}{-2.5pt}
		\begin{tabular}{|c|c|c|c|}
			\hline
			 Dataset & \# of Training Samples & \# of Test Samples & Resolution \\\hline\hline
			M2014 \cite{scharstein2014high} & 15 & 15 & 2960$\times$1942 \\\hline
% 			K2012 & 194 & 195 & 1242$\times$375 \\\hline
			K2015 \cite{kitti2015} & 200 & 200 & 1242$\times$375 \\\hline
			ETH3D \cite{eth3d2017} & 27 & 20 & 960$\times$480 \\\hline
			SF \cite{mayer2016large} & 35454 & 4370 & 960$\times$540 \\\hline
		\end{tabular}
\end{table}
The distribution of disparity of different datasets is quite different, which is an important factor to guide the network design, especially the disparity search range in the point-wise correlation layer discussed in Section \ref{subsec:pwc}. We statistic the disparity distribution from the ground truth of the above datasets as shown in Fig.~\ref{fig:distribution}.

\textbf{Baselines}. We choose existing state-of-the-art DNNs in estimating disparity from stereo images. In terms of ED-Conv2D, we choose DispNetC \cite{mayer2016large}, CRL \cite{crl2017}, DN-CSS \cite{flownet3}, AnyNet \cite{anynet2019}, and FADNet~\cite{wang2020fadnet}. Regarding CVM-Conv3D, we use PSMNet~\cite{psmnet2018}, GANet \cite{ganet2019}, GWCNet \cite{gwcnet2019}, AANet \cite{xu2020aanet}, and LEAStereo \cite{cheng2020hierarchical}. From the model accuracy's perspective, GANet and LEAStereo are the main top-ranked methods, while from the inference performance's perspective, AnyNet and FADNet are very efficient. Comparing with these baselines, we will show how our new proposed framework balance the model accuracy and inference speed.

\textbf{Implementation Details.}
We firstly pre-train FADNet++ on the SceneFlow training samples for 90 epochs. Following the finetuning strategy proposed in \cite{cfnet2021}, we then jointly finetune our pre-trained FADNet++ on the combination of training samples in M2014, K2015 and ETH3D for another 2400 epochs. 

\subsection{Model Accuracy}
In this subsection, we train the chosen models on the selected datasets and evaluate their model accuracy (EPE, endpoint error). We follow the same training scheme \cite{cfnet2021} that first trains a base model on the SceneFlow dataset, and fine-tunes the model on other datasets.

\begin{table}[!ht]
	\centering
	\caption{Model accuracy on the SceneFlow dataset. Bold indicates the best. Underline indicates the second best. The runtime is the inference time measured on an Nvidia Tesla V100.}
	\label{table:sf_results_epe}
	\addtolength{\tabcolsep}{-3.5pt}
	\begin{tabular}{|c|c|c|c|c|} \hline
	{Type} &	{Method} & \makecell{GPU Memory \\ Footprint [GB]} & {EPE [px]} & {Runtime [s]} \\ \hline\hline
	\multirow{4}{*}{ED-Conv2D} & DispNetC \cite{mayer2016large} & 1.9 & 1.68 & 0.015 \\
		&   CRL \cite{crl2017} & 2.2 & 1.32 & 0.026 \\
% 		&   DN-CSS \cite{flownet3} & 109.02 & (0.78) & 0.037 \\
		&   AnyNet \cite{anynet2019} & \textbf{1.31} & 3.39 & \textbf{0.013} \\ 
		&   FADNet \cite{wang2020fadnet} & 2.6 & 0.83 & 0.048 \\ \hline
	\multirow{5}{*}{CVM-Conv3D} &	PSMNet\cite{psmnet2018} & 5.6 & 1.09 & 0.619 \\
		&   GANet\cite{ganet2019} & 7.5 & 0.78 & 2.292 \\
		&   GWCNet\cite{gwcnet2019} & 5.7 & \underline{0.77} & 0.260 \\
		&   AANet\cite{xu2020aanet} & 1.91 & 0.87 & 0.086 \\
		&   LEAStereo\cite{cheng2020hierarchical} & 25.3 & 0.78 & 0.478 \\ \hline
	\multirowcell{4}{Configurable\\ ED-Conv2D} &   FADNet++ & 2.3 & \textbf{0.76} & 0.033 \\
		&   FADNet-M & \underline{1.7} & 0.91 & 0.019 \\
		&   FADNet-S & 1.8 & 1.19 & 0.015 \\ 
		&   FADNet-T & 1.9 & 1.83 & \underline{0.014} \\ \hline
	\end{tabular}
\end{table}

\textbf{SceneFlow.} The accuracy comparison of different models is shown in Table~\ref{table:sf_results_epe}. In terms of EPE on the SceneFlow dataset, we can see that our FADNet++ outperforms all the other models including both ED-Conv2D and CVM-Conv3D, which shows the capability of our model to capture the disparity information of stereo images.

Compared to ED-Conv2D methods, our FADNet++ significantly improves the model accuracy with comparable inference time. For example, in ED-Conv2D, the best accuracy model is FADNet with EPE of 0.83, whose inference time is 0.048 seconds. Our FADNet++ outperforms FADNet in both EPE (with around 9\% improvement) and runtime (with around 50\% faster speed). In terms of the runtime of ED-Conv2D, AnyNet is very efficient with on 0.013 seconds, but its EPE is very high, which is far away from real-world production. Our configurable feature of FADNet++ enables to configure different sizes of models to balance EPE and runtime. For example, FADNet-T is as efficient as AnyNet, but it achieves around 80\% lower EPE than AnyNet. With a larger model of our FADNet-M, the runtime is only 0.003 longer than AnyNet, but our method can achieve 3.7 times lower EPE than AnyNet.

Compared to CVM-Conv3D methods, our FADNet++ achieves better EPE and inference time. Existing GANet, GWCNet, and LEAStereo obtain about 0.77-0.78 EPE on SceneFlow with more than 0.27 inference time, while our FADNet++ achieves 0.76 EPE with a magnitude smaller inference time. Even the very efficient 3D mode of AANet, it runs at 0.07 seconds, which is more than 2 times slower than FADNet++, and its EPE is still larger than ours.

We also analyze the GPU memory footprint needed to support the runtime execution of each network. The memory space is typically used to hold the model parameters, the optimizer status and the intermediate output tensors \cite{rasley2020deepspeed}. The memory footprint is managed by the deep learning toolkit, such as PyTorch in our implementation, and related to not only the network characteristics listed above but also the chosen network forwarding/back-propagation algorithms and the memory caching scheme. Notice that the CVM-Conv3D methods usually suffer from large memory requirements and fail to be deployed on those low-end computing devices. However, our FADNet++ and its variants only consume nearly 2 GB of memory space, which make them feasible in many platforms. We also observe that FADNet-S and FADNet-T consumes a bit more memory space than FADNet-M. The reason is that the cuDNN library may choose different convolution algorithms, which consume different sizes of memory, for different layer channel settings to achieve the best model inference efficiency. 

\begin{figure*}[!ht]
    \captionsetup[subfigure]{labelformat=empty}
	\centering
	\subfloat[]
	{
	\adjincludegraphics[width=0.23\linewidth,trim={{.25\width} {.18\width} {.35\width} {.12\width}},clip]{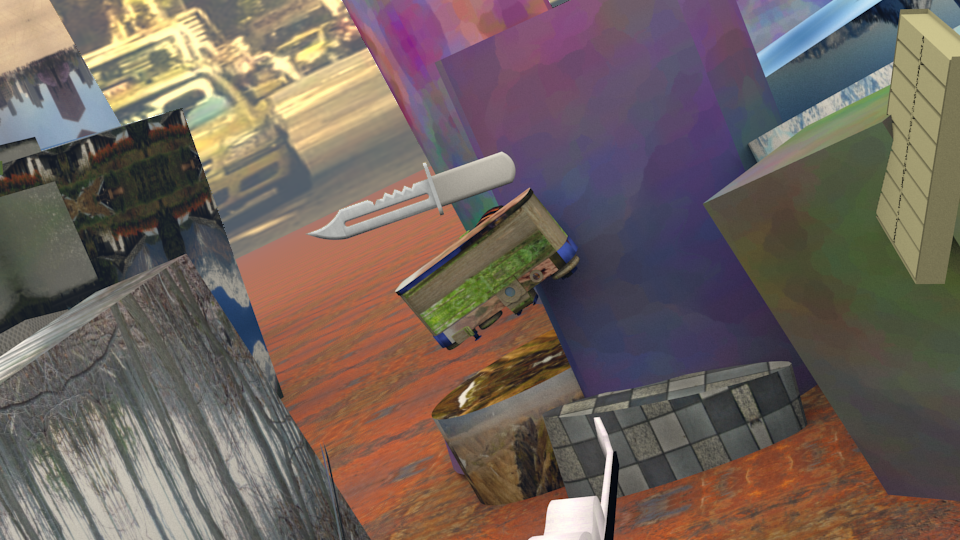}\label{fig:vt_bf_left_rgb}
	}
	\subfloat[]
	{
	\adjincludegraphics[width=0.23\linewidth,trim={{.25\width} {.18\width} {.35\width} {.12\width}},clip]{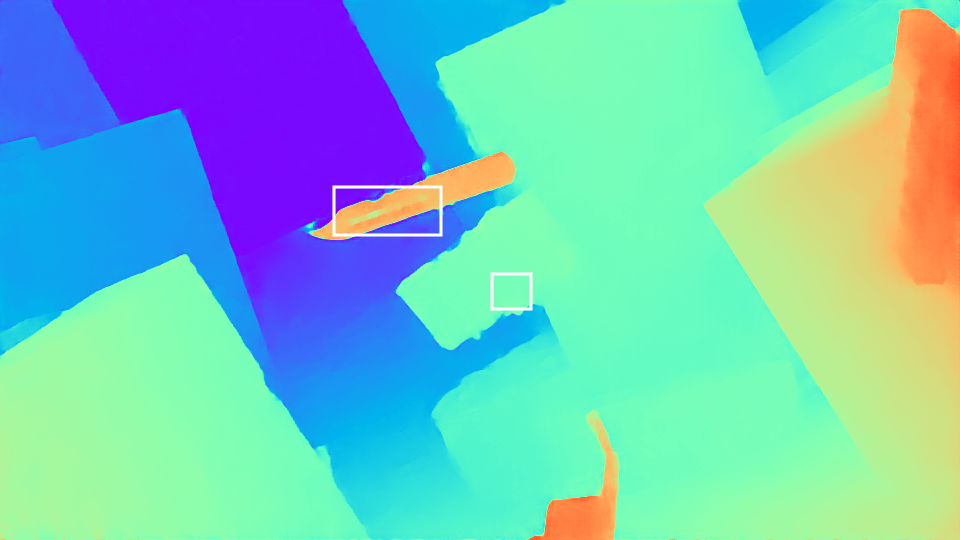}\label{fig:vt_bf_disp_gt}
	} % D1_all: 2.66\%
	\subfloat[]
	{
	\adjincludegraphics[width=0.23\linewidth,trim={{.25\width} {.18\width} {.35\width} {.12\width}},clip]{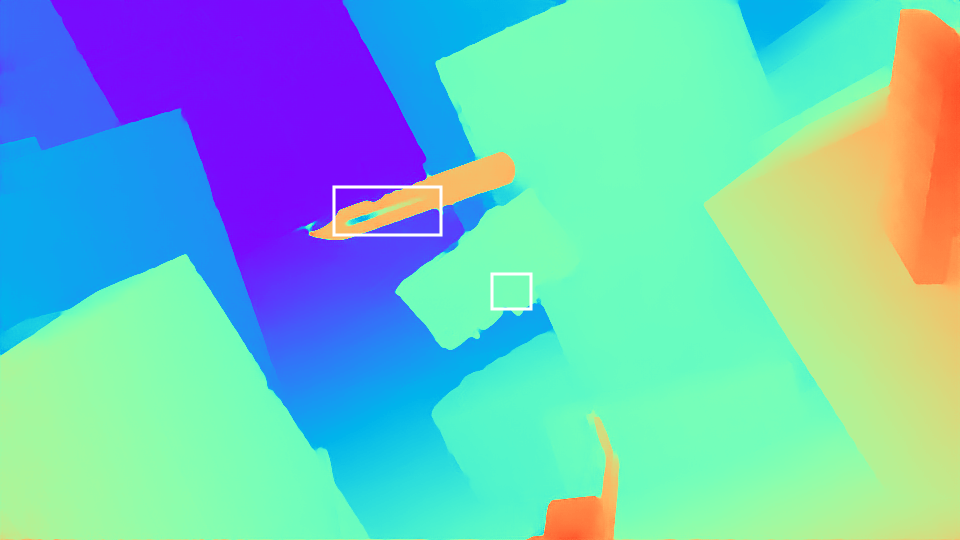}\label{fig:vt_bf_disp_gt} % D1_all: 1.35\%
	}
	\subfloat[]
	{
	\adjincludegraphics[width=0.23\linewidth,trim={{.25\width} {.18\width} {.35\width} {.12\width}},clip]{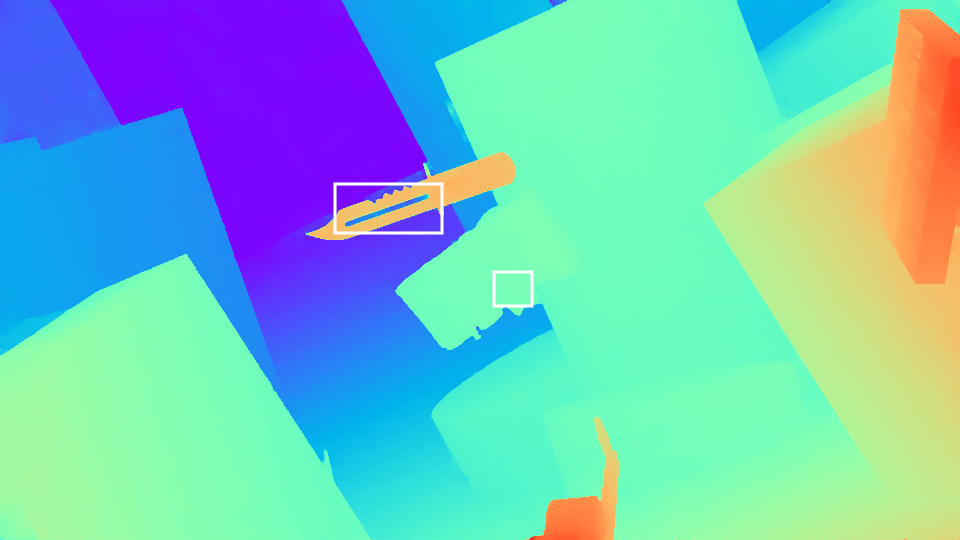}\label{fig:vt_bf_disp_gt} % D1_all: 2.02\%
	}
 	\vspace{-2.0 em}
 	\qquad
	\subfloat[(a) Left/Right image]
	{
	\adjincludegraphics[width=0.23\linewidth,trim={{.25\width} {.18\width} {.35\width} {.12\width}},clip]{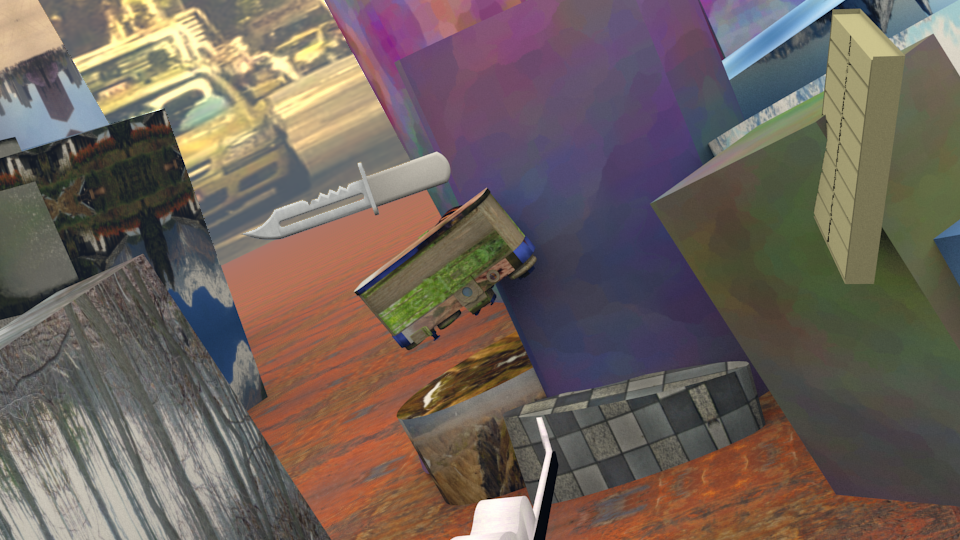}\label{fig:vt_bf_left_rgb}
	}
	\subfloat[(b) DispNetC (0.02 s) \cite{mayer2016large}]
	{
	\adjincludegraphics[width=0.23\linewidth,trim={{.25\width} {.18\width} {.35\width} {.12\width}},clip]{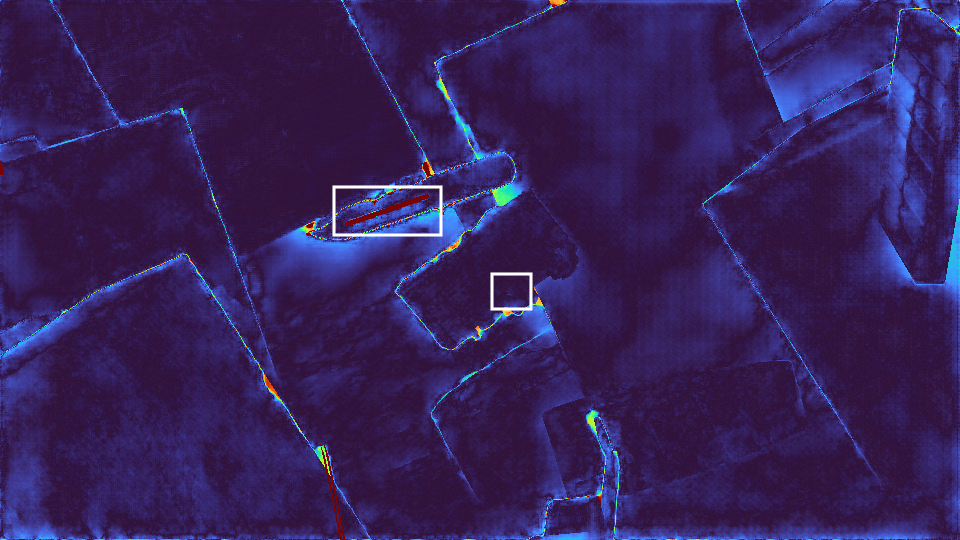}\label{fig:vt_bf_disp_gt}
	}
	\subfloat[(c) CRL (0.03 s) \cite{crl2017}]
	{
	\adjincludegraphics[width=0.23\linewidth,trim={{.25\width} {.18\width} {.35\width} {.12\width}},clip]{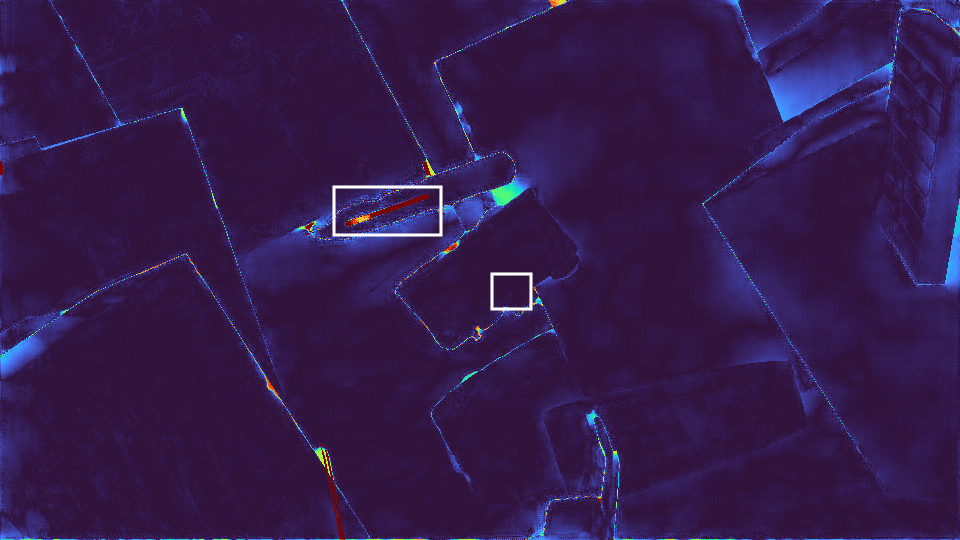}\label{fig:vt_bf_disp_gt}
	}
	\subfloat[(d) AANet (0.09 s) \cite{xu2020aanet}]
	{
	\adjincludegraphics[width=0.23\linewidth,trim={{.25\width} {.18\width} {.35\width} {.12\width}},clip]{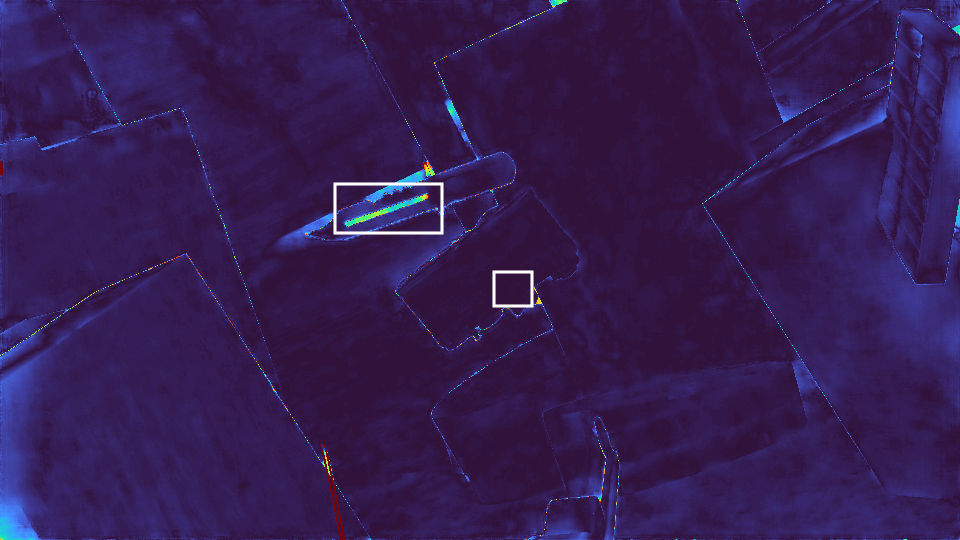}\label{fig:vt_bf_disp_gt}
	}
	\qquad
	\subfloat[]
	{
	\adjincludegraphics[width=0.23\linewidth,trim={{.25\width} {.18\width} {.35\width} {.12\width}},clip]{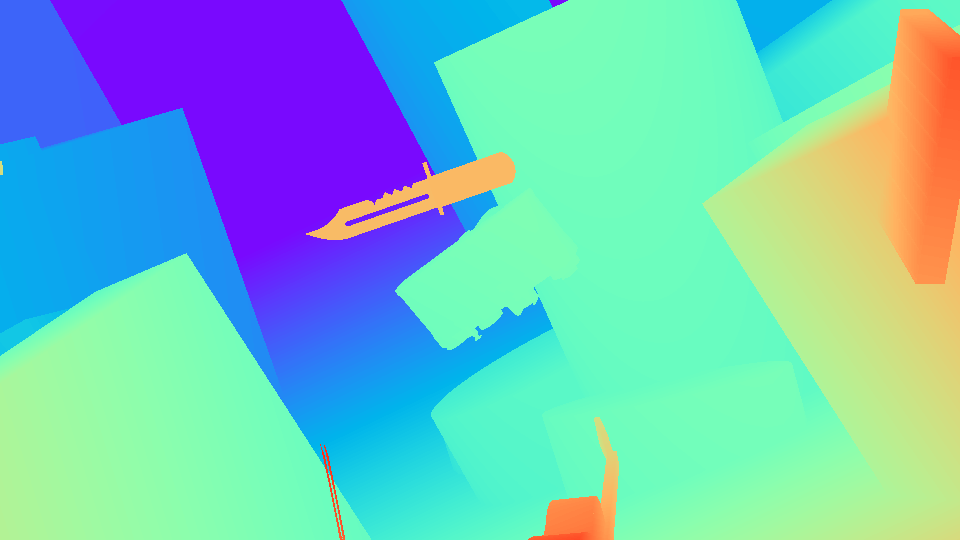}\label{fig:vt_bf_left_rgb}
	}
	\subfloat[]
	{
	\adjincludegraphics[width=0.23\linewidth,trim={{.25\width} {.18\width} {.35\width} {.12\width}},clip]{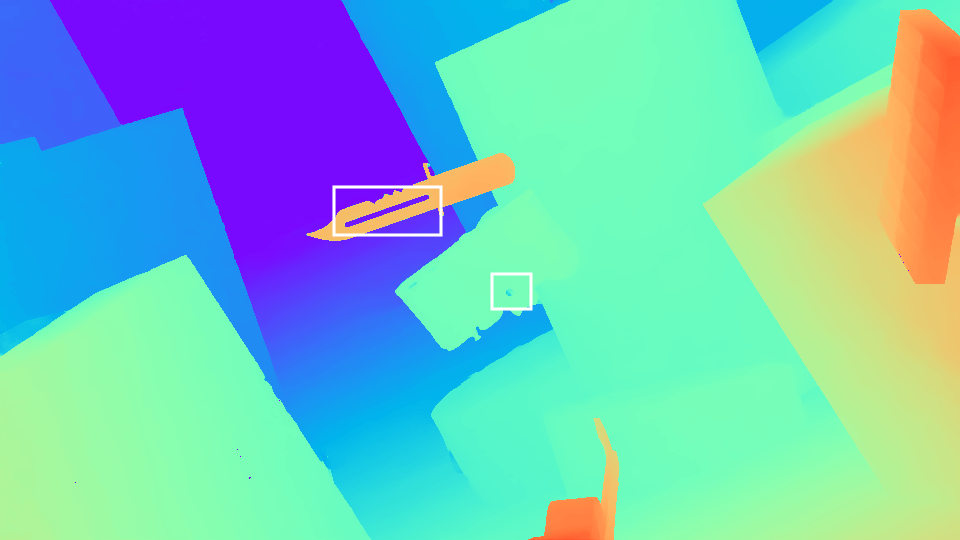}\label{fig:vt_bf_disp_gt}
	} % D1_all: 2.66\%
	\subfloat[]
	{
	\adjincludegraphics[width=0.23\linewidth,trim={{.25\width} {.18\width} {.35\width} {.12\width}},clip]{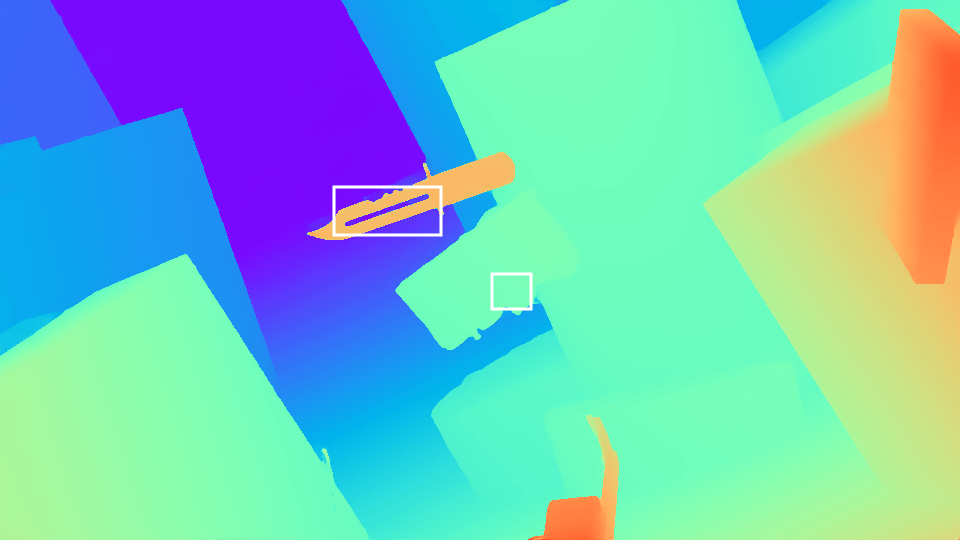}\label{fig:vt_bf_disp_gt} % D1_all: 1.35\%
	}
	\subfloat[]
	{
	\adjincludegraphics[width=0.23\linewidth,trim={{.25\width} {.18\width} {.35\width} {.12\width}},clip]{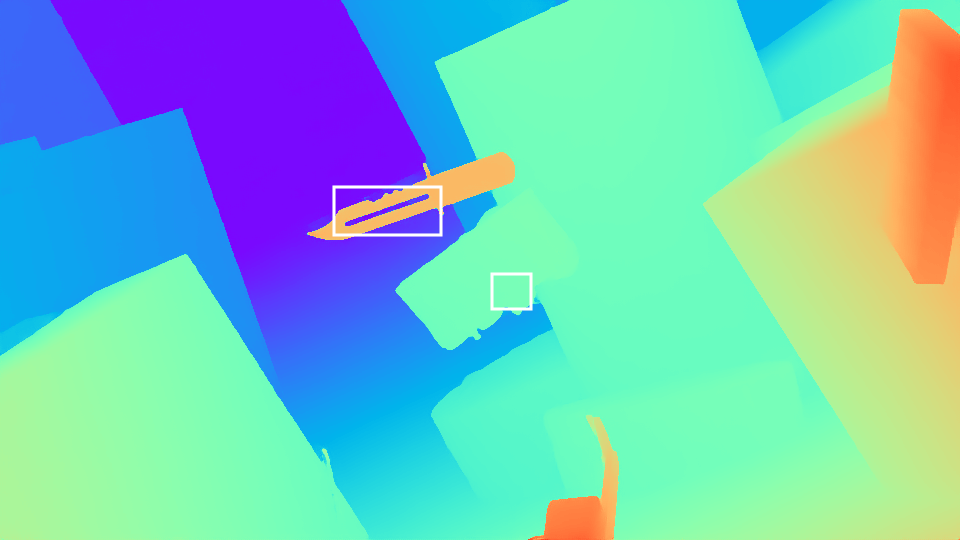}\label{fig:vt_bf_disp_gt} % D1_all: 2.02\%
	}
 	\vspace{-2.0 em}
 	\qquad
	\subfloat[(e) Groundtruth]
	{
	\adjincludegraphics[width=0.23\linewidth,trim={{.25\width} {.18\width} {.35\width} {.12\width}},clip]{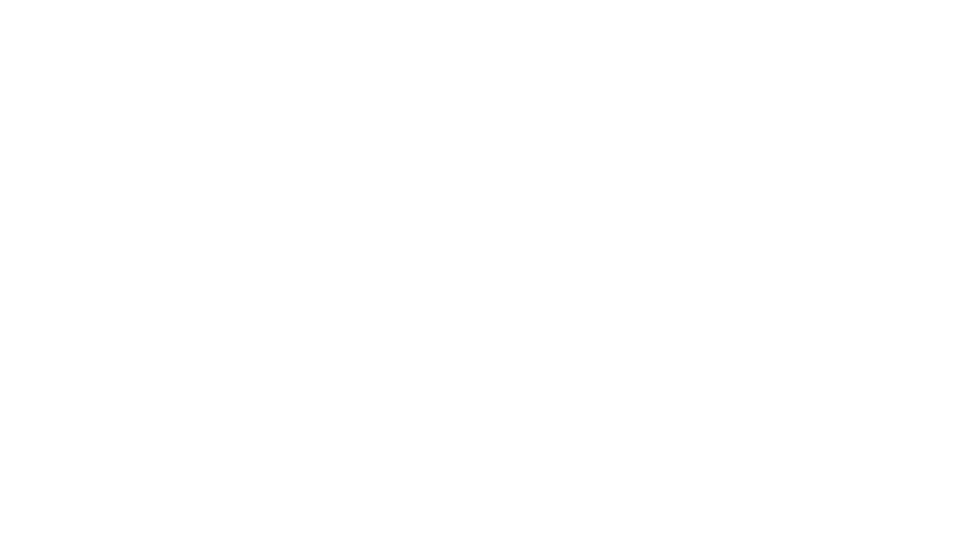}\label{fig:vt_bf_left_rgb}
	}
	\subfloat[(f) GANet (2.29 s) \cite{ganet2019}]
	{
	\adjincludegraphics[width=0.23\linewidth,trim={{.25\width} {.18\width} {.35\width} {.12\width}},clip]{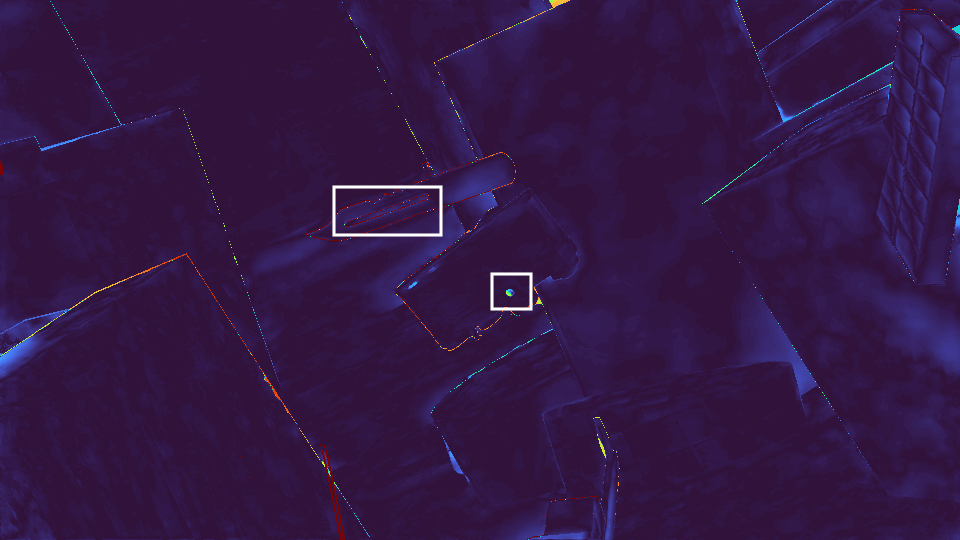}\label{fig:vt_bf_disp_gt}
	}
	\subfloat[(g) LEAStereo (0.48 s) \cite{cheng2020hierarchical}]
	{
	\adjincludegraphics[width=0.23\linewidth,trim={{.25\width} {.18\width} {.35\width} {.12\width}},clip]{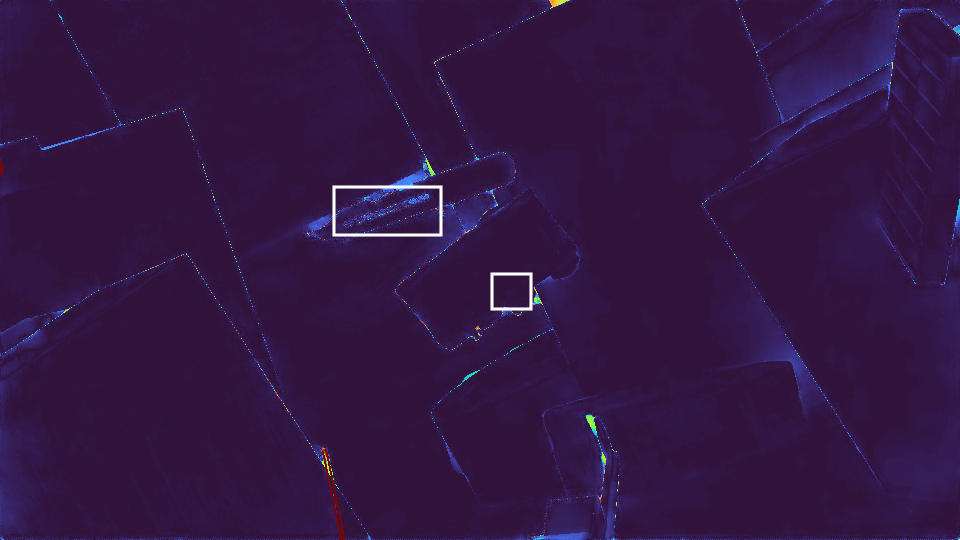}\label{fig:vt_bf_disp_gt}
	}
	\subfloat[(h) Our FADNet++ (0.03 s)]
	{
	\adjincludegraphics[width=0.23\linewidth,trim={{.25\width} {.18\width} {.35\width} {.12\width}},clip]{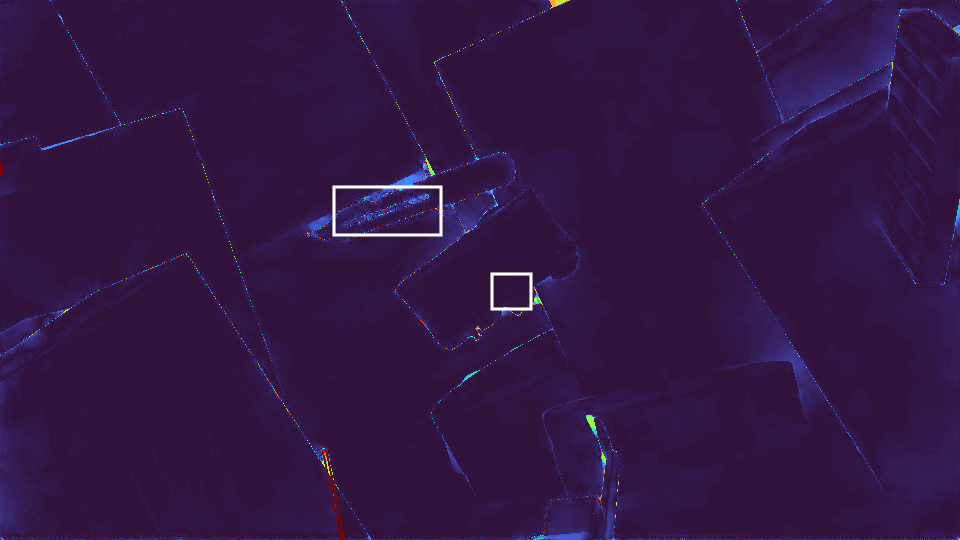}\label{fig:vt_bf_disp_gt}
	}
	\caption{An illustration of the disparities produced by different methods. The sample is from SceneFlow and has a resolution of 960$\times$540 The input left and right images and the groundtruth disparity are shown in (a) and (e), respectively. (b)-(d) and (f)-(h) show the predicted disparity maps of different methods as well as their error maps. The colder color in error maps indicate lower errors. The parentheses in the sub-captions include the runtime of each method on an Nvidia Tesla V100.}
	\label{fig:vis_results_on_sf}
\end{figure*}

The visualization of some samples is shown in Fig.~\ref{fig:vis_results_on_sf}, which compares our FADNet++ with two ED-Conv2D networks, DispNetC and CRL, and three CVM-Conv3D networks, AANet, GANet and LEAStereo. It is observed that DispNetC and CRL fail to produce accurate disparities for the object boundaries. Besides, the hole of the knife cannot be correctly recognized by those two ED-Conv2D methods. On the contrary, our FADNet++ can work well on the boundaries and the details of the knife. The predicted disparity map of FADNet++ is close to those of AANet, GANet and LEAStereo, while FADNet++ runs much faster than those CVM-Conv3D methods.

\textbf{Robotics Vision Challenge.}
To demonstrate the model robustness on different scenarios, we utilize the similar strategy as \cite{cfnet2021}, where we validate our model on three realistic stereo datasets using the Robotics Vision Challenge (RVC) 2020\footnote{\url{http://www.robustvision.net/index.php}}. In RVC, each model is required to be trained in the dataset combined with M2014, K2015 and ETH3D, and it then is evaluated on M2014, K2015 and ETH3D separately. We choose top-ranked representative models (i.e., from top-1 to top-6, the models are CFNet~\cite{cfnet2021}, NLCANet\_V2~\cite{nlcanet2020}, HSMNet~\cite{hsmnet2019}, CVANet\footnote{CVANet has no published paper and code, so we cannot evaluate its runtime.}, AANet~\cite{xu2020aanet}, and GANet~\cite{ganet2019}, respectively) on the RVC leaderboard\footnote{\url{http://www.robustvision.net/leaderboard.php?benchmark=stereo}} to compare the model accuracy and the inference speed. 

\begin{table*}[!ht]
	\centering
	\caption{Joint generalization comparison on RVC with ETH3D, Middlebury, and KITTI2015 datasets. Rank indicates the ranking of model accuracy on the RVC leaderboard. Bold indicates the best. Underline indicates the second best. The runtime is measured with the input resolution (1242$\times$375) of the KITTI2015 dataset, and other resolutions should have similar patterns.}
	\label{tab:rvc_results}
	\begin{tabular}{|c|c|c|c|c|c|c|c|c|c|c|c|} \hline
		\multirow{2}{*}{Method} & \multicolumn{3}{c|}{KITTI2015} & \multicolumn{3}{c|}{Middlebury2014} & \multicolumn{3}{c|}{ETH3D2017} & Runtime & \multirow{2}{*}{Rank} \\ \cline{2-10}
		& D1\_bg & D1\_fg & D1\_all & bad 4.0 & rms & avg error & bad 1.0 & bad 2.0 & avg error & [s] & \\ \hline\hline
		CFNet \cite{cfnet2021} & \underline{1.65} & \underline{3.53} & \underline{1.96} & 11.3 & \underline{18.2} & \underline{5.07} & \textbf{3.7} & \textbf{0.97} & \textbf{0.26} & 0.234 & 1\\ 
		NLCANet\_V2 \cite{nlcanet2020} & \textbf{1.51} & 3.97 & \textbf{1.92} & \underline{10.3} & 21.9 & 5.60 & \underline{4.11} & \underline{1.2} & 0.29 & 0.44 & 2\\ 
		HSMNet \cite{hsmnet2019} & 2.74 & 8.73 & 3.74 & \textbf{9.68} & \textbf{13.4} & \textbf{3.44} & 4.40 & 1.51 & \underline{0.28} & 0.15 & 3\\ 
		CVANet & 1.74 & 4.98 & 2.28 & 23.1 & 25.9 & 8.64 & 4.68 & 1.37 & 0.34 & - & 4 \\
		AANet \cite{xu2020aanet} &  2.23 & 4.89 & 2.67 & 25.8 & 32.8 & 12.8 & 5.41 & 1.95 & 0.33 & \underline{0.062} & 5\\
		GANet \cite{ganet2019} & 1.88 & 4.58 & 2.33 & 16.3 & 42.0 & 15.8 & 6.97 & 1.25 & 0.45 & 1.71 & 6 \\ \hline
		FADNet++ & 1.99 & \textbf{3.18} & 2.19 & 31.4 & 27.7 & 11.9 & 4.36 & 1.30 & 0.34 & \textbf{0.029} & - \\ \hline
		%FADNet++\_RVC & 2.62/2.48/1.99 & \underline{3.68}/4.10/3.18 & 2.80/2.75/2.19 & 47.7/63.6 & 33.0/45.6 & 21.0/11.9 & 4.24/4.11/4.36 & 1.32/1.28/1.30 & 0.31/0.31/0.34 \\ \hline
	\end{tabular}
\end{table*}

The results are shown in Table~\ref{tab:rvc_results}. The runtime for different models is measured on the same platform using their open-sourced code to guarantee fair comparison. The runtime in CVANet is empty as it has no publicly available code and paper. It can be seen that the performance of our model is ranked from 3-5 in the three datasets among the top-6 models. Specifically, in KITTI2015, our model is slightly worse than CFNet and NLCANet\_V2, and it outperforms other four models in terms of the metric of D1\_all. In the average error of the M2014 dataset, our FADNet++ still outperforms AANet and GANet. Regarding the ETH3D dataset, our model outperforms GANet and is comparable with CVANet and GANet. In summary, the top-3 models have good model accuracy, but their inference time is very slow, while our FADNet++ achieves a magnitude order of faster speed. Compared with the top-4 to top-6 models, FADNet++ achieves comparable model accuracy while achieving around $3\times$ faster than AANet and around $59\times$ faster than GANet. Note that among the compared methods, only our FADNet++ can provide real-time inference speed (i.e., $\geq 30$FPS) on a Tesla V100 GPU.

\begin{figure*}[!ht]
    %\captionsetup[subfigure]{labelformat=empty, farskip=0pt}
	\centering
	\subfloat[Left image]
	{
	\includegraphics[width=0.23\linewidth]{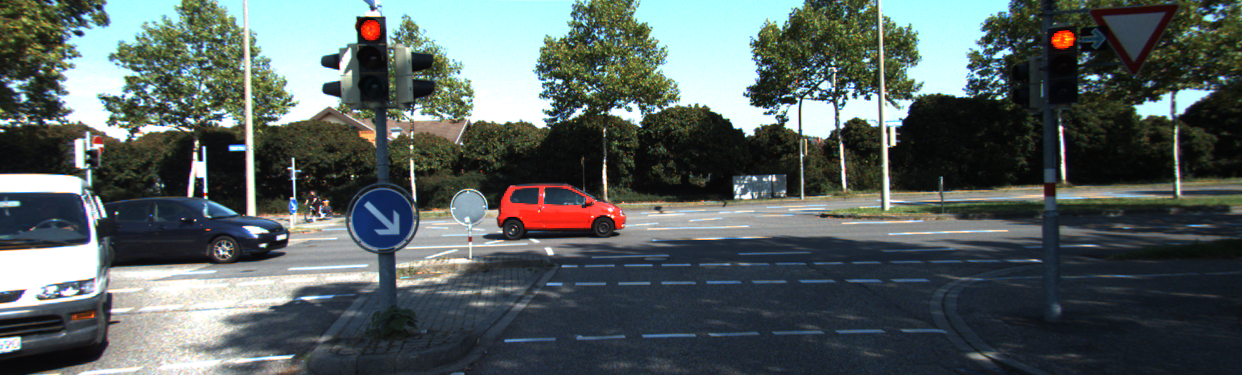}\label{fig:vt_bf_left_rgb}
	}
	\subfloat[GANet \cite{ganet2019}]
	{
	\includegraphics[width=0.23\linewidth]{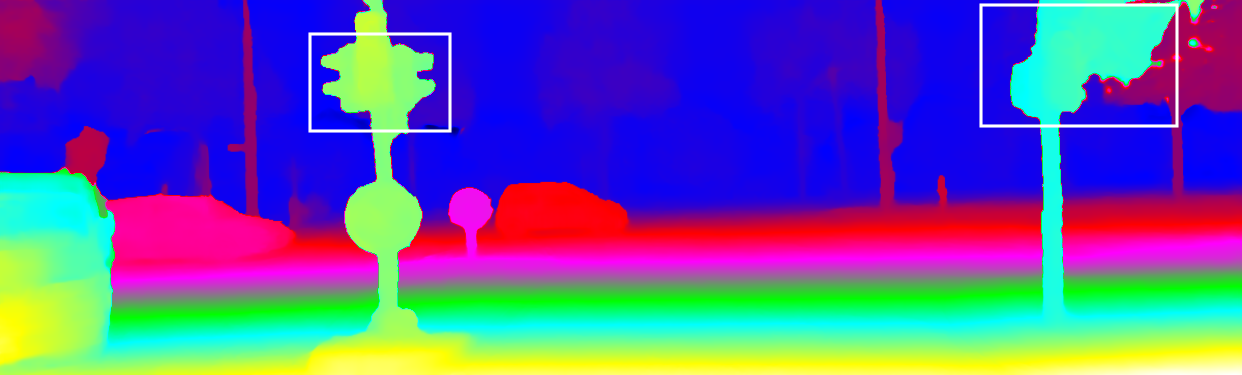}\label{fig:vt_bf_disp_gt}
	} % D1_all: 2.66\%
	\subfloat[AANet \cite{xu2020aanet}]
	{
	\includegraphics[width=0.23\linewidth]{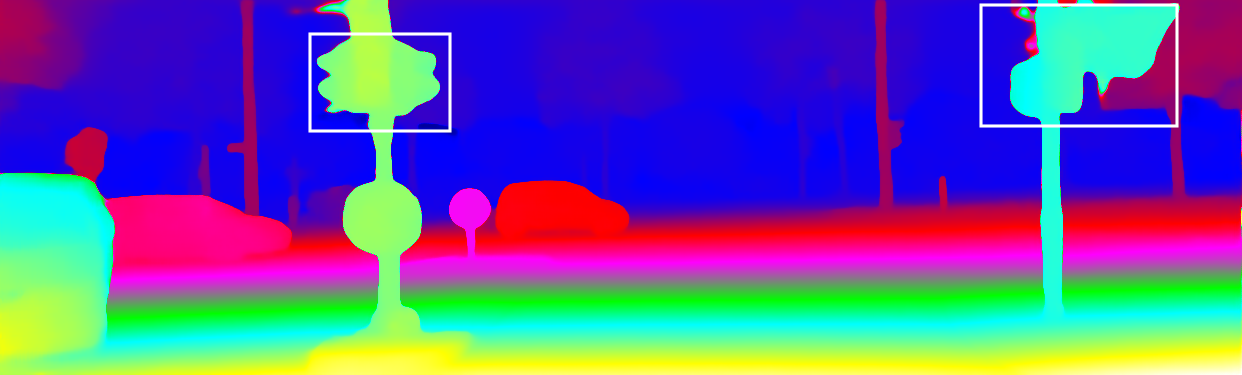}\label{fig:vt_bf_disp_gt} % D1_all: 1.35\%
	}
	\subfloat[HSMNet \cite{hsmnet2019}]
	{
	\includegraphics[width=0.23\linewidth]{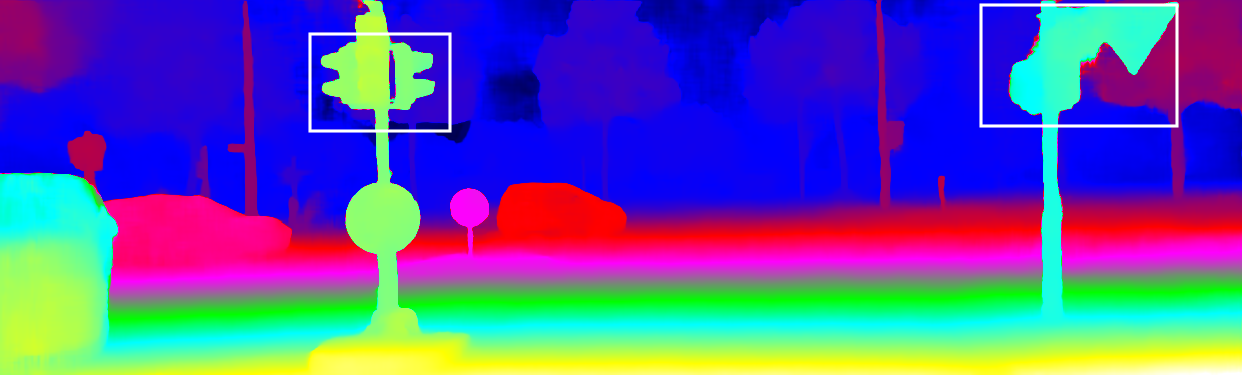}\label{fig:vt_bf_disp_gt} % D1_all: 2.02\%
	}
 	\qquad
	\subfloat[Right image]
	{
	\includegraphics[width=0.23\linewidth]{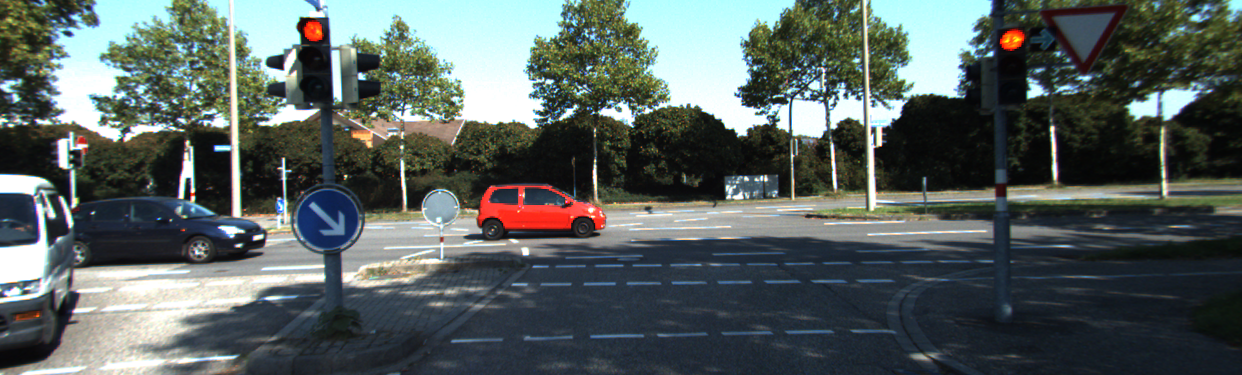}\label{fig:vt_bf_left_rgb}
	}
	\subfloat[NLCANet \cite{nlcanet2020}]
	{
	\includegraphics[width=0.23\linewidth]{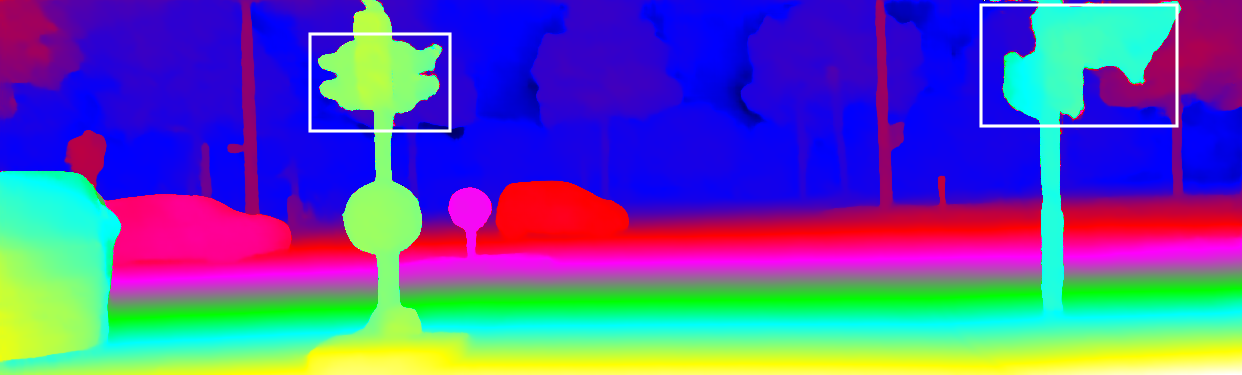}\label{fig:vt_bf_disp_gt}
	}
	\subfloat[CFNet \cite{cfnet2021}]
	{
	\includegraphics[width=0.23\linewidth]{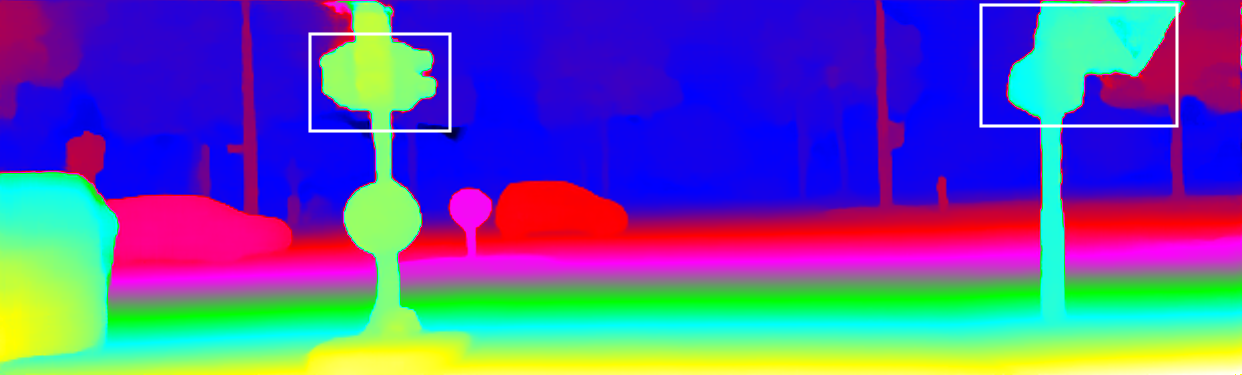}\label{fig:vt_bf_disp_gt}
	}
	\subfloat[FADNet++ (Ours)]
	{
	\includegraphics[width=0.23\linewidth]{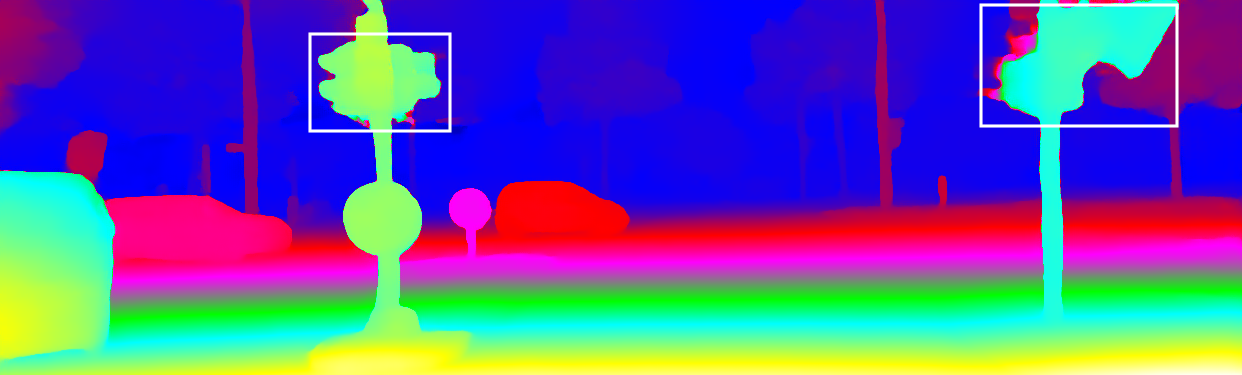}\label{fig:vt_bf_disp_gt}
	}
	\caption{Results achieved on the KITTI 2015 dataset. }
	\label{fig:results_on_k2015}
\end{figure*}

\begin{figure*}[ht]
    %\captionsetup[subfigure]{labelformat=empty, farskip=0pt}
	\centering
	\subfloat[Left image]
	{
	\includegraphics[width=0.23\linewidth]{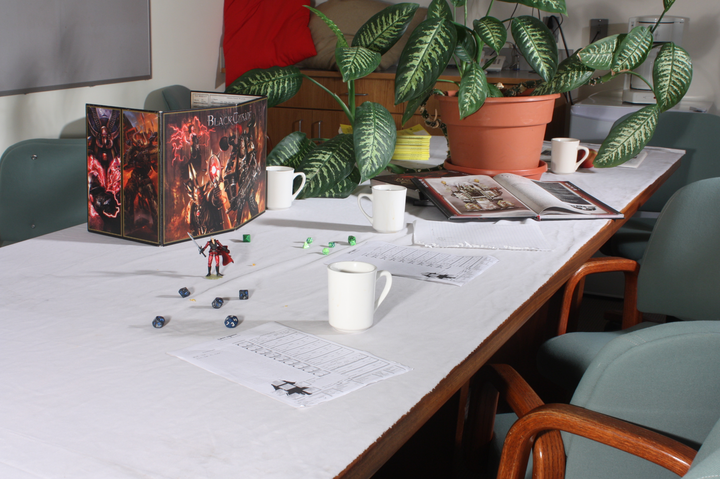}\label{fig:vt_bf_left_rgb}
	}
	\subfloat[GANet \cite{ganet2019}]
	{
	\includegraphics[width=0.23\linewidth]{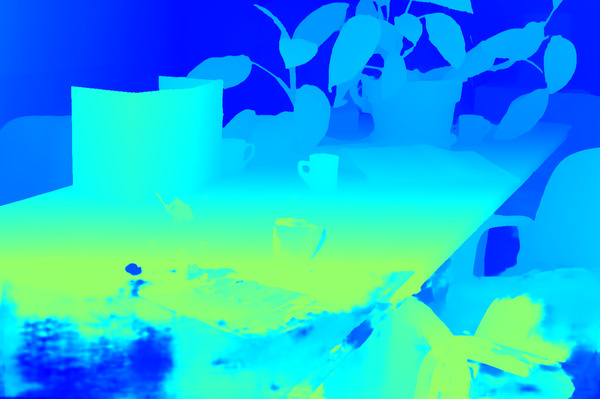}\label{fig:vt_bf_disp_gt}
	} % D1_all: 2.66\%
	\subfloat[AANet \cite{xu2020aanet}]
	{
	\includegraphics[width=0.23\linewidth]{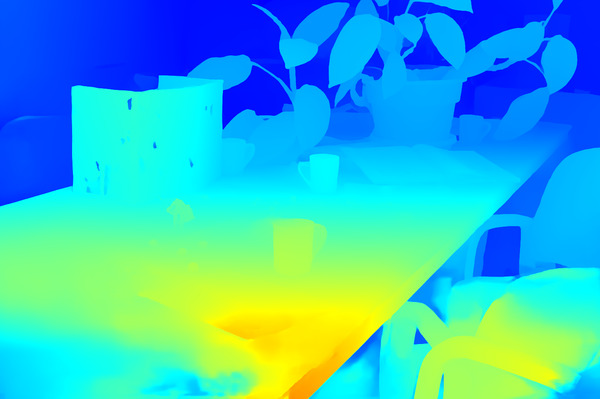}\label{fig:vt_bf_disp_gt} % D1_all: 1.35\%
	}
	\subfloat[HSMNet \cite{hsmnet2019}]
	{
	\includegraphics[width=0.23\linewidth]{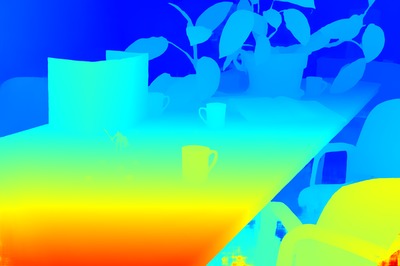}\label{fig:vt_bf_disp_gt} % D1_all: 2.02\%
	}
 	\qquad
	\subfloat[Right image]
	{
	\includegraphics[width=0.23\linewidth]{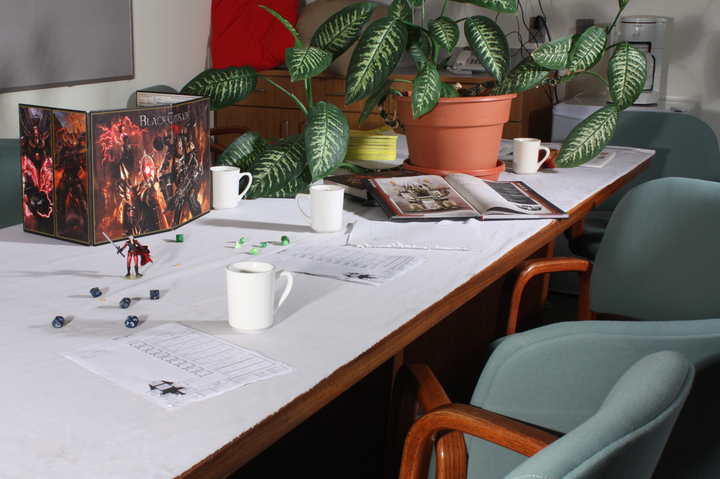}\label{fig:vt_bf_left_rgb}
	}
	\subfloat[NLCANet \cite{nlcanet2020}]
	{
	\includegraphics[width=0.23\linewidth]{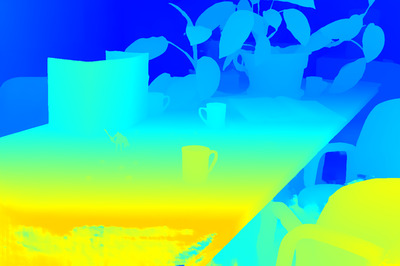}\label{fig:vt_bf_disp_gt}
	}
	\subfloat[CFNet \cite{cfnet2021}]
	{
	\includegraphics[width=0.23\linewidth]{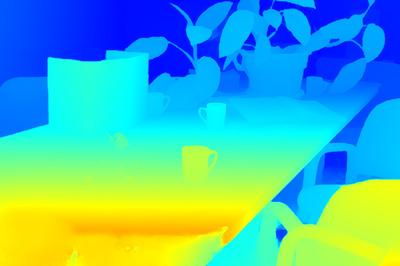}\label{fig:vt_bf_disp_gt}
	}
	\subfloat[FADNet++ (Ours)]
	{
	\includegraphics[width=0.23\linewidth]{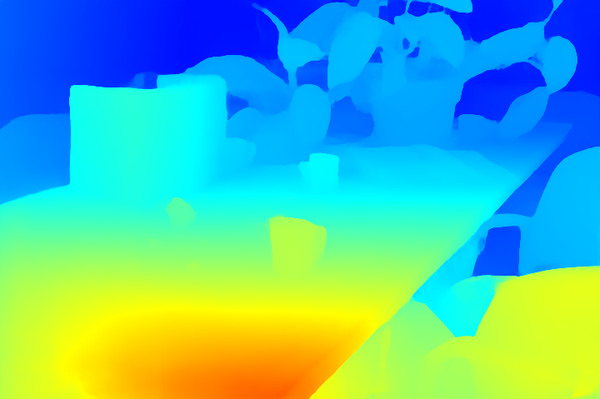}\label{fig:vt_bf_disp_gt}
	}
	\caption{Results achieved on the Middlebury 2014 test set. The image pair show above is taken from the CrusadeP data. Our method generates smooth results close to HSMNet and CFNet and performs better than GANet and AANet, especially for the white flat desk.}
	\label{fig:results_on_md}
\end{figure*}

\begin{figure}[ht]
	\centering
	\subfloat[Left image]
	{
	\includegraphics[width=0.47\linewidth]{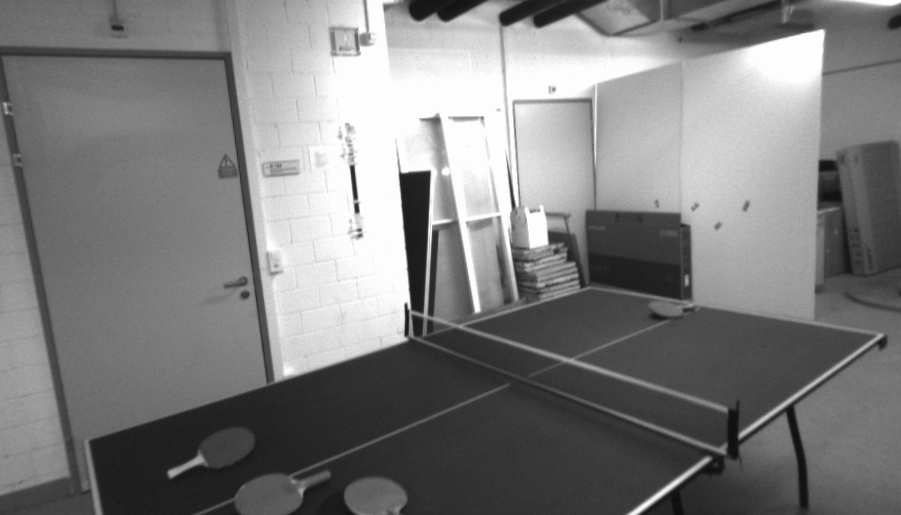}\label{fig:vt_bf_left_rgb}
	}
    \subfloat[Right image]
	{
	\includegraphics[width=0.47\linewidth]{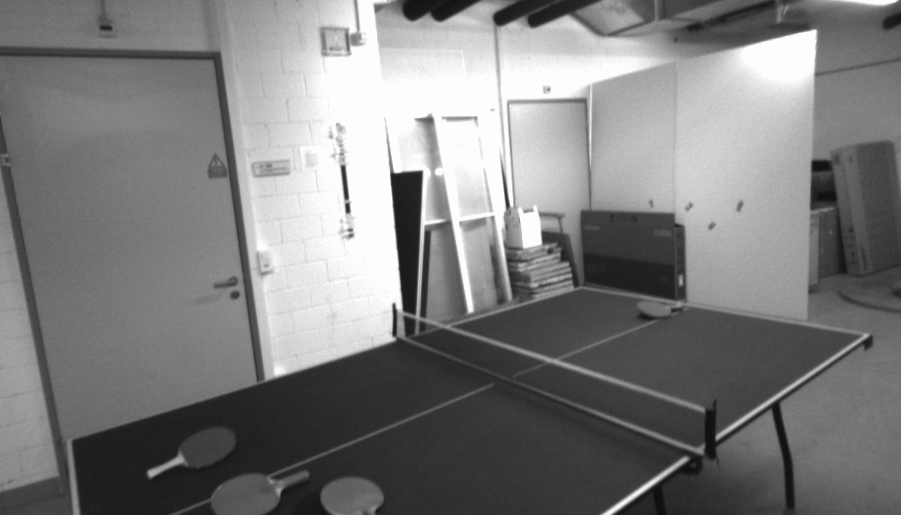}\label{fig:vt_bf_left_rgb}
	} 
	\qquad
	\subfloat[CBMV \cite{cbmv2018}]
	{
	\includegraphics[width=0.47\linewidth]{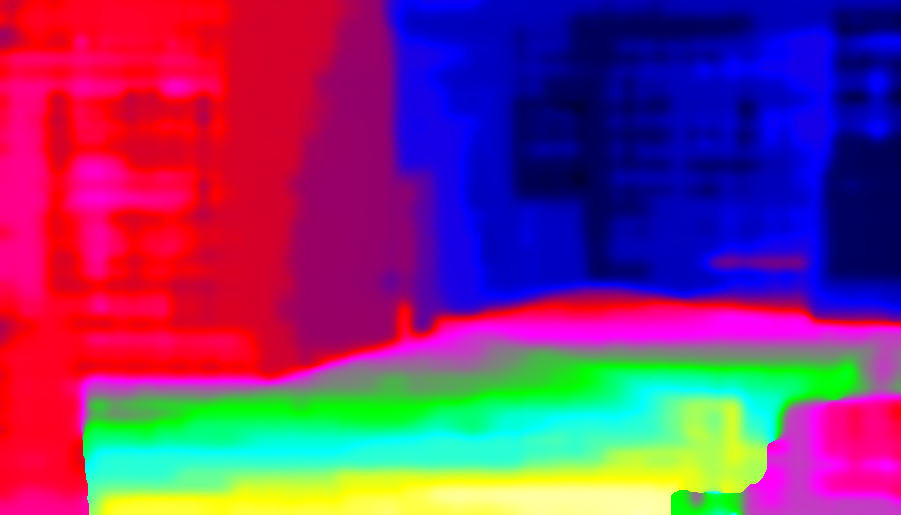}\label{fig:vt_bf_left_rgb}
	}
    \subfloat[SGM-Forest \cite{sgmforest2018}]
	{
	\includegraphics[width=0.47\linewidth]{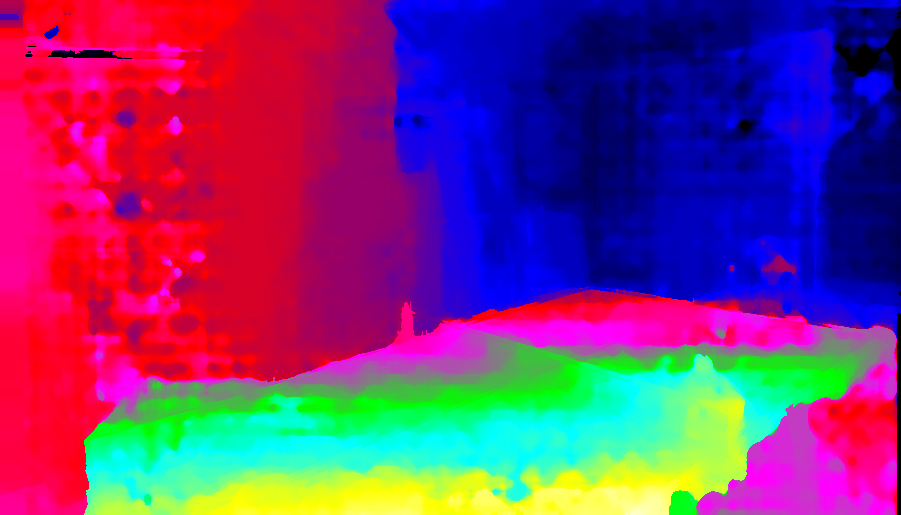}\label{fig:vt_bf_left_rgb}
	} 
	\qquad
	\subfloat[CFNet \cite{cfnet2021}]
	{
	\includegraphics[width=0.47\linewidth]{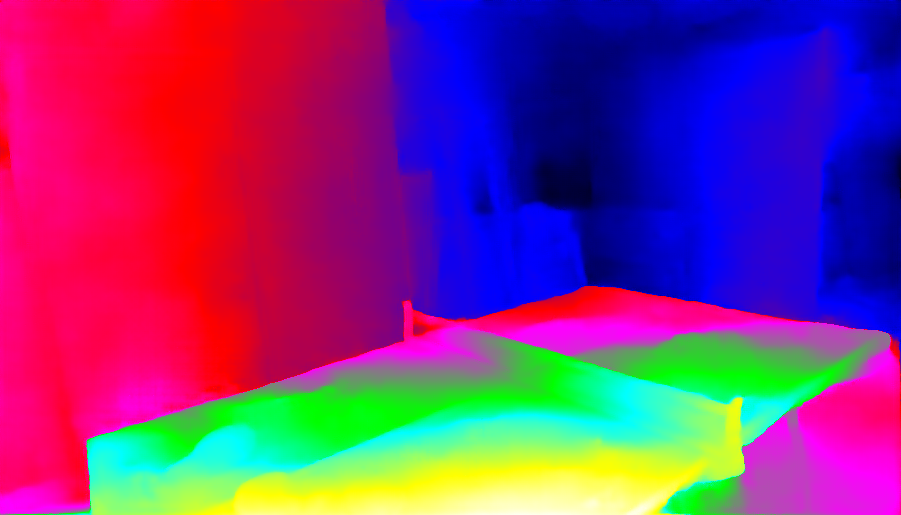}\label{fig:vt_bf_left_rgb}
	}
    \subfloat[FADNet++ (ours)]
	{
	\includegraphics[width=0.47\linewidth]{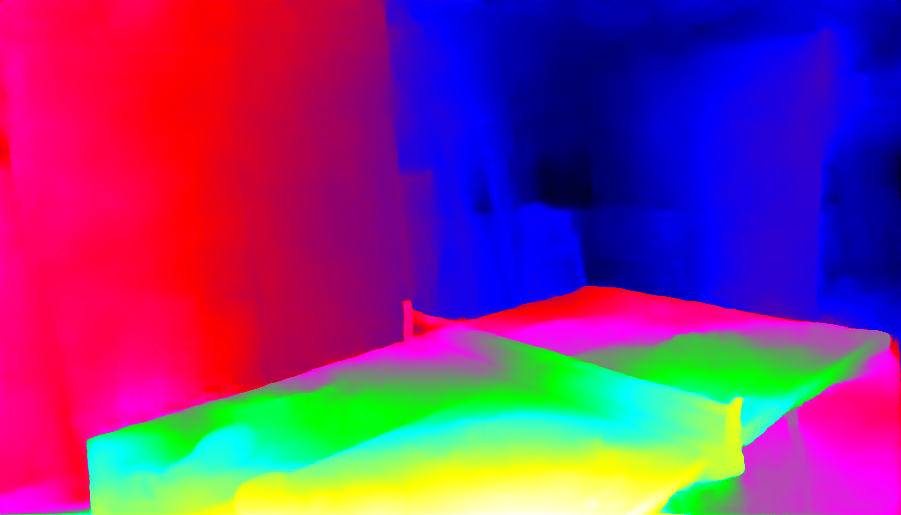}\label{fig:vt_bf_left_rgb}
	} 
	\caption{Results achieved on the ETH3D 2017 test set. The storage\_room\_2\_2l image pair is used for test. The disparity map generated by FADNet++ is close to the Top-1 CFNet in RVC2020, and much smoother than two traditional SOTA stereo matching methods, especially on the ping-pong table.}
	\label{fig:results_on_eth3d}
\end{figure}

Some visualization effects on K2015, M2014, and ETH3D datasets are shown in Fig.~\ref{fig:results_on_k2015}, Fig.~\ref{fig:results_on_md}, and Fig.~\ref{fig:results_on_eth3d} respectively. 
For K2015, compared to GANet and AANet, our FADNet++ can generate disparity maps with richer details (see left white boxes in Fig. \ref{fig:results_on_k2015}) and smoother results (see right white boxes in Fig. \ref{fig:results_on_k2015}).
For M2014, from the white desk in Fig. \ref{fig:results_on_md}, it can also be clearly observed that our method produces much better and smoother results than GANet and AANet.
For ETH3D, it is clear that our FADNet++ performs well on textureless regions (such as the ping-pong table). Its disparity is close to the top-1 CFNet and much smoother than those achieved by other traditional SOTA methods.

\subsection{Inference Efficiency}
In the above subsection, we have shown that our model achieves comparable model accuracy while providing very efficient inference speed on the Tesla V100 GPU. In this subsection, we provide more experimental results on inference GPUs and mobile GPUs to show how our configurable model achieves real-time inference performance on different platforms with good model accuracy.

\begin{table}[!ht]
	\centering
	\caption{Quantitative results on the SceneFlow dataset among different inference servers. Bold indicates the best. Underline indicates the second best. }
	
	\label{tab:sf_results_server_gpus}
	\begin{tabular}{|c|c|c|c|c|} \hline
		\multirow{2}{*}{Method} & \multirow{2}{*}{EPE [px]} & \multicolumn{3}{c|}{Runtime [s]} \\ \cline{3-5}
		& & RTX2070 & P40 & T4 \\ \hline\hline
		DispNetC\cite{mayer2016large} & 1.68 & 0.022 & 0.025 & 0.04 \\
		CRL\cite{crl2017}     & 1.32 & 0.042 & 0.047 & 0.074 \\
% 		DN-CSS\cite{flownet3} & (0.78) & 0.063 & 0.069 & 0.107\\
		AnyNet\cite{anynet2019}  & 3.39 & \textbf{0.012} & \textbf{0.017} & \textbf{0.014} \\ 
		FADNet\cite{wang2020fadnet} & 0.83 & 0.085 & 0.096 & 0.146 \\\hline
		PSMNet\cite{psmnet2018} & 1.09 & 0.571 & 0.492 & 0.792 \\
		GANet\cite{ganet2019} & 0.78 & 5.2 & 5.5 & 7.344 \\
		GWCNet\cite{gwcnet2019} & \underline{0.77} & 0.45 & 0.421 & 0.646 \\
		AANet\cite{xu2020aanet}  & 0.87 & 0.124 & 0.183 & 0.23 \\
		LEAStereo\cite{cheng2020hierarchical} & 0.78 & 0.851 & 0.71 & 0.978 \\\hline
		FADNet++ & \textbf{0.76} & 0.053 & 0.06 & 0.091 \\  
		FADNet-M & 0.91 & 0.025 & 0.031 & 0.037 \\
		FADNet-S & 1.19 & 0.017 & 0.023 & 0.023 \\ 
		FADNet-T & 1.83 & \underline{0.013} & \underline{0.02} & \underline{0.015} \\ \hline
	\end{tabular}
\end{table}
\textbf{On Inference Server GPUs.} The inference performance on the inference servers is shown in Table~\ref{tab:sf_results_server_gpus}. In terms of the runtime, we can see that AnyNet achieves the fastest speed among the evaluated methods, but its EPE on the SceneFlow dataset is extremely high (3.39). Our FADNet-T achieves very close inference speed with AnyNet while achieving around $83\%$ improvement in EPE. Being aimed to achieving real-time inference speed (i.e., $\geq$30FPS whose inference time should be around 0.033s), our FADNet-M can provide real-time inference speed in all three inference server GPUs with the EPE of 0.91. The other existing model, DispNetC, who also achieves real-time inference speed in all inference servers, has the EPE of 1.68, which is around $87\%$ higher than ours. Even the CVM-Conv3D based models achieve very good model accuracy, they run very slow on these inference GPUs so that they are far away from production to provide real-time disparity estimation. In summary, our configuration framework can be configured as a relatively small model (i.e., FADNet-M) compared to FADNet++ and provides real-time inference speed with good model accuracy.

\begin{table}[!ht]
	\centering
	\caption{Quantitative results on SceneFlow dataset on among different mobile platforms. Bold indicates the best. Underline indicates the second best. }
	\label{tab:sf_results_mobile_gpus}
	\begin{tabular}{|c|c|c|c|c|} \hline
		\multirow{2}{*}{Method} & \multirow{2}{*}{\makecell{GPU Memory \\ Footprint [GB]}} & \multirow{2}{*}{EPE [px]} & \multicolumn{2}{c|}{Runtime [s]} \\ \cline{4-5}
		& & & TX2 & AGX \\ \hline\hline
		DispNetC\cite{mayer2016large} & 3.9 & 1.68 & 0.309 & 0.108 \\
		StereoNet\cite{stereonet2018} & 9.5 & 1.10 & 1.148 & 0.282 \\
		AnyNet\cite{anynet2019} & \textbf{3.1} & 3.39 & \underline{0.125} & \textbf{0.041} \\ 
		AANet\cite{xu2020aanet} & 12.6 & 0.87 & 1.83 & 0.585 \\
		FADNet\cite{wang2020fadnet} & 4.9 & \underline{0.83} & 1.176 &  0.413 \\ \hline
		FADNet++ & 4.3 & \textbf{0.76} & 0.735 & 0.258 \\
		FADNet-M & \underline{3.7} & 0.91 & 0.335 & 0.113\\
		FADNet-S & 3.8 & 1.19 & 0.192 & 0.068 \\ 
		FADNet-T & 3.9 & 1.83 & \textbf{0.111} & \underline{0.043}  \\ \hline
	\end{tabular}
\end{table}
\textbf{On Mobile GPUs.} 
To demonstrate the feasibility of our model applying on mobile devices, we choose two model GPUs (Nvidia TX2 and AGX) to compare the performance. Due to the memory limitation, all the CVM-Conv3D methods cannot run on such mobile devices. Therefore, we compare the inference speed with ED-Conv2D methods and also include the occupied GPU memory footprints. The results are shown in Table~\ref{tab:sf_results_mobile_gpus}. Again, AnyNet still has very fast inference speed even on mobile GPUs, but its EPE is rather high. Our configured model FADNet-T achieves very close inference speeds with AnyNet while it has much better model accuracy than AnyNet. Comparing between our configured FANet-S and StereoNet, both of which have similar model accuracy (EPE is around 1.1-1.2), we can see that FADNet-S runs $4\times$ and $5.9\times$ faster than StereoNet on TX2 and AGX GPUs, respectively. In summary, our configurable framework enables us to set different sizes of models for adapting on different computing power devices with reasonable model accuracy. We also profile the device memory usage of different models. Notice that there are no CVM-Conv3D models since they fail to run on our tested mobile platforms due to the memory limitation. Compared to the existing real-time networks like DispNetC and AnyNet, our FADNet++ and FADNet-M achieve much lower EPEs with similar memory usage. Besides, since the cuDNN library in PyTorch may use different convolution algorithms for different layer channel numbers to achieve the best inference speed, it is possible that the smaller FADNet-S and FADNet-T can even consume a bit large memory than FADNet-M. In addition, the GPU memory usage of the same network can be also different between two computing platforms, such as 2.3 GB on V100 but 4.3 GB on AGX for FADNet++. On the one hand, the memory space on Jetson TX and AGX is shared by both the CPU and GPU so that the memory management strategy is different from the pure GPU memory on V100. On the other hand, the cuDNN library may also have different implementations for the X86-based and ARM-based systems, respectively. 

We put our configured models on FADNet++ running on all evaluated devices in Table~\ref{tab:all_results_fadnet}, which shows the configurable feature of our model for balancing model accuracy and inference speeds on different hardware.

\begin{table}[!ht]
	\centering
	\caption{Configurable speed vs. model accuracy (EPE on the SceneFlow dataset) on different GPUs. }
	\addtolength{\tabcolsep}{-1.5pt}
	\label{tab:all_results_fadnet}
	\begin{tabular}{|c|c|c|c|c|c|c|c|} \hline
		\multirow{2}{*}{Model} & EPE & \multicolumn{6}{c|}{Runtime [s]} \\\cline{3-8} 
		& [px] & RTX2070 & P40 & T4 & V100 & TX2 & AGX \\ \hline\hline
		FADNet++ & 0.76 & 0.053 & 0.06  & 0.091 & 0.032 & 0.735 & 0.258 \\
		FADNet-M & 0.91 & 0.025 & 0.031 & 0.037 & 0.016 & 0.335 & 0.113\\
		FADNet-S & 1.19 & 0.017 & 0.023 & 0.023 & 0.015 & 0.192 & 0.068 \\ 
		FADNet-T & 1.83 & 0.013 & 0.02  & 0.015 & 0.013 & 0.111 & 0.043  \\ \hline
	\end{tabular}
\end{table}

\section{Conclusion} \label{sec:conclusion}
In this paper, we proposed an efficient yet accurate neural network, FADNet++, for end-to-end disparity estimation to embrace the time efficiency and estimation accuracy on the stereo matching problem. The proposed FADNet++ exploits point-wise correlation layers, residual blocks, and multi-scale residual learning strategy to make the model be accurate in many scenarios while preserving fast inference time. Moreover, to adapt to the target computing devices of different capability, we design a simple but effective configurable channel scaling ratio that can generate various FADNet++ variants of different inference performance. Our training solution can be applied to all the variants and boost their highest accuracy. We conducted extensive experiments to compare our FADNet++ with existing state-of-the-art 2D and 3D based methods in terms of accuracy and speed. 
%Our experiments cover four popular stereo datasets and a wide range of computing GPUs from server-level to edge-level. 
Experimental results showed that FADNet++ achieves comparable accuracy while it runs much faster than the 3D based models. Compared to the existing mobile solution, FADNet++ achieves a competitive inference speed of 15 FPS with nearly three times accurate. 

We have two future directions following our discovery in this paper. First, we would like to improve the disparity estimation accuracy on the low-end devices. To approach the accuracy of FADNet++ produced by the server GPUs, it is necessary to explore the techniques of model compression, including pruning, quantization, and so on. Second, we would also like to apply AutoML \cite{he2019automl} for searching a well-performing network structure for disparity estimation.

% \appendices
% \section{Proof of the First Zonklar Equation}
% Appendix one text goes here.

\section*{Acknowledgments}
This research was supported by Hong Kong RGC GRF
grant HKBU 12200418. We thank the anonymous reviewers for their constructive comments and suggestions. We would also like to thank NVIDIA AI Technology Centre (NVAITC) for providing the GPU clusters for some experiments.

% Can use something like this to put references on a page
% by themselves when using endfloat and the captionsoff option.
\ifCLASSOPTIONcaptionsoff
  \newpage
\fi

\bibliographystyle{IEEEtran}
\bibliography{main.bbl}

\begin{IEEEbiography}
[{\includegraphics[width=1in,height=1.25in,clip,keepaspectratio]
{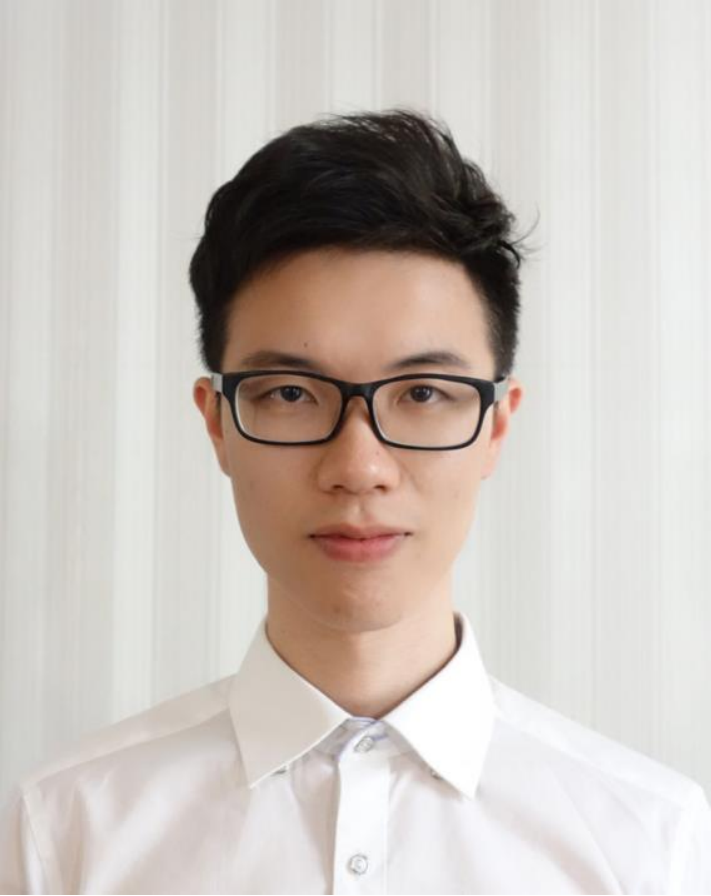}}]{Qiang Wang}
is currently a Research Assistant Professor at Department of Computer Science, Hong Kong Baptist University. He received his B.Sc. degree from South China University of Technology in 2014, and the PhD degree at Department of Computer Science, Hong Kong Baptist University. His research interests include General-Purpose GPU Computing, Deep Learning and 3D Vision. He is a recipient of Hong Kong PhD Fellowship.
\end{IEEEbiography}

\begin{IEEEbiography}
	[{\includegraphics[width=1in,height=1.25in,clip,keepaspectratio]
		{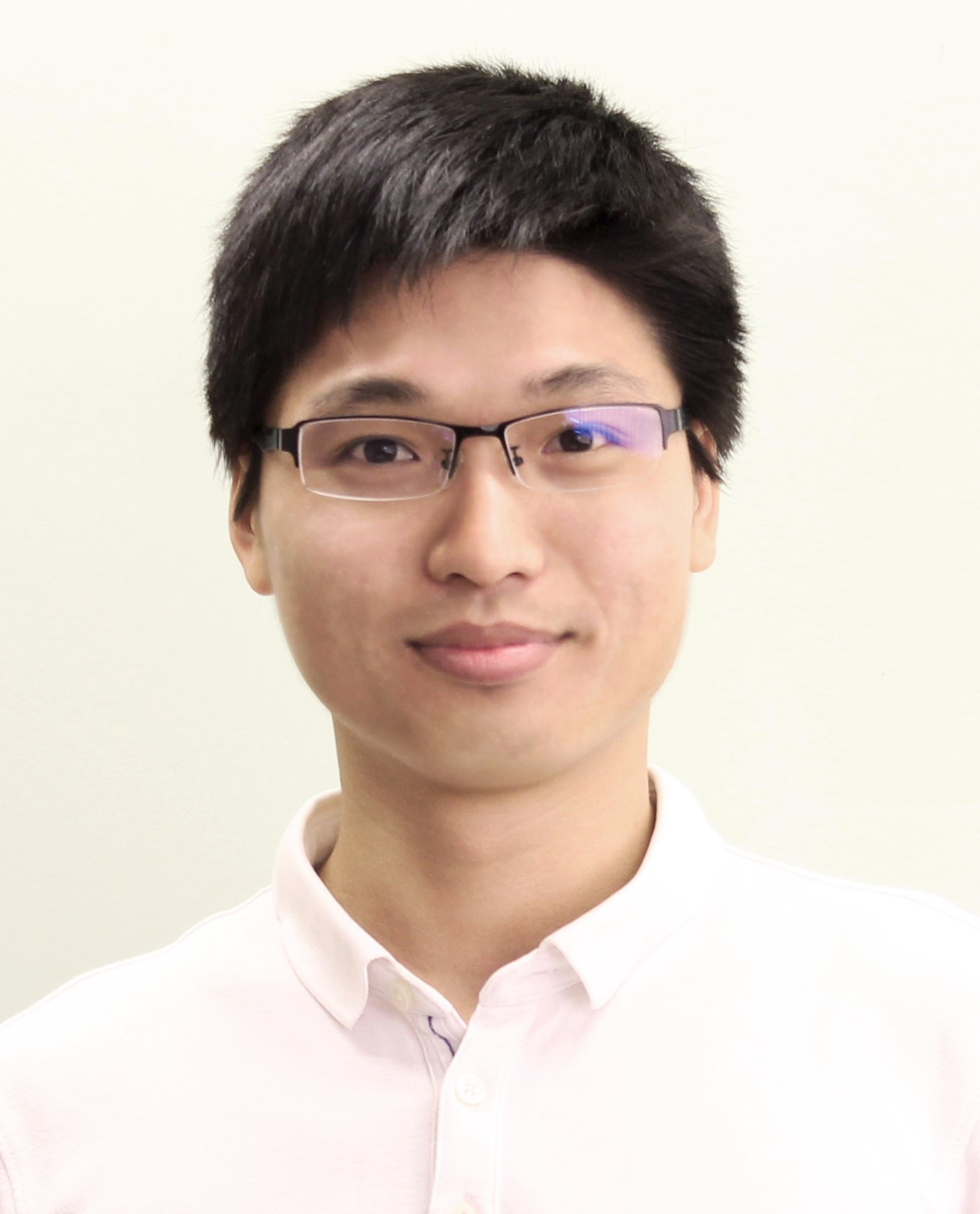}}]{Shaohuai Shi}
	is currently a Research Assistant Professor with the Department of Computer Science and Engineering, Hong Kong University of Science and Technology. He received the BE degree in software engineering from the South China University of Technology, P.R. China, in 2010, the MS degree in computer science from the Harbin Institute of Technology, P.R. China, in 2013, and the PhD degree in computer science from Hong Kong Baptist University, in 2020. His research interests include GPU computing and machine learning systems.
\end{IEEEbiography}

\begin{IEEEbiography}
	[{\includegraphics[width=1in,height=1.25in,clip,keepaspectratio]
		{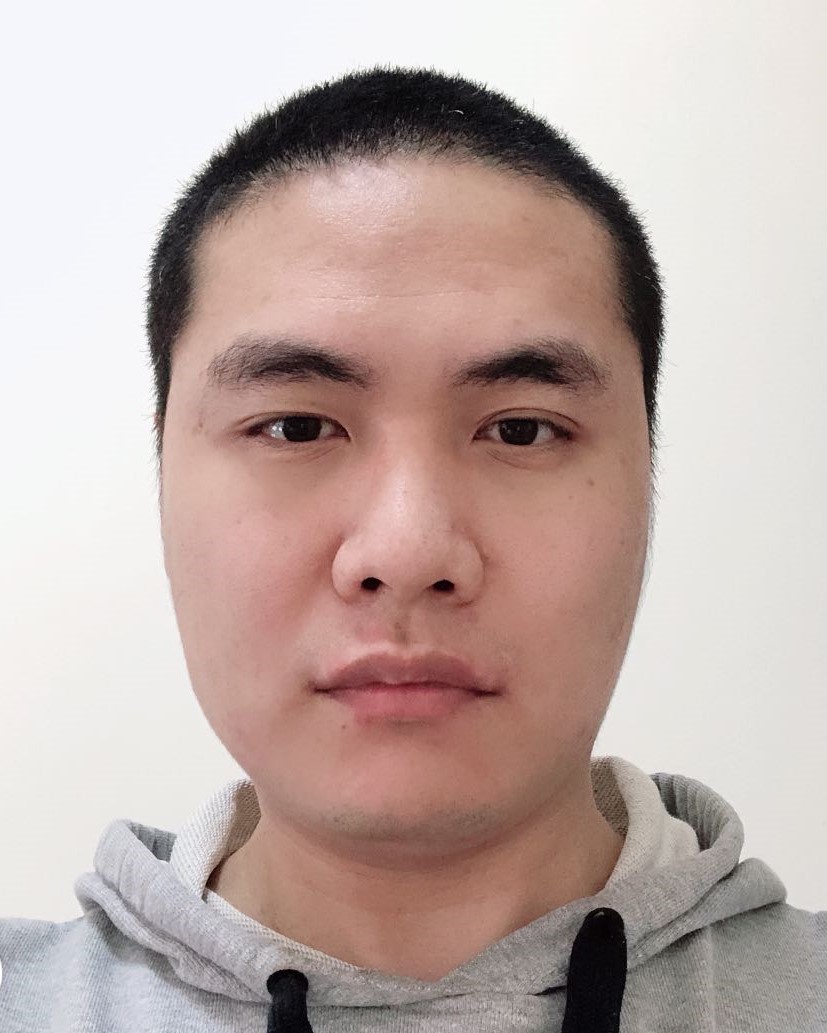}}]{Shizhen Zheng}
	received his B.Sc. degree from Sun Yat-sen University in 2012. Currently, he is a Senior Software Engineer in ParallelDomain Inc.. He used to be a Senior Research Assistant at the Department of Computer Science, Hong Kong Baptist University. His research interests include Deep Learning, 3D Vision, Synthetic Data and Simulator.
\end{IEEEbiography}

\begin{IEEEbiography}
	[{\includegraphics[width=1.15in,height=1.35in,clip,keepaspectratio]
		{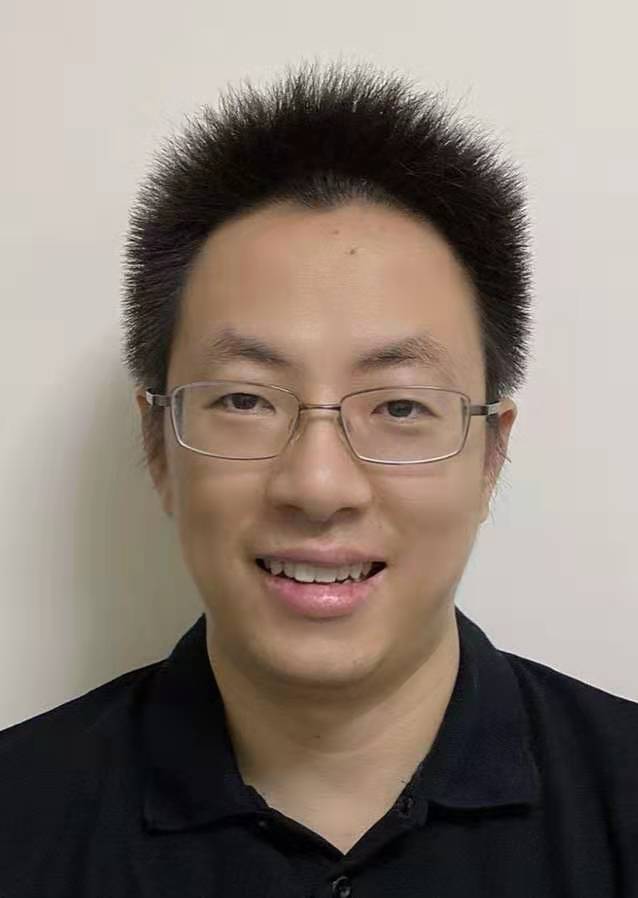}}]{Kaiyong Zhao}
	received the B.E. degree in computer science from Beijing Institute of Technology, P.R. China, in 2005, the MS degree and the Ph.D. degree in computer science from Department of Computer Science, Hong Kong Baptist University in 2011 and 2016, respectively. His research interests include distributed and parallel computing, GPU computing, SLAM and 3D reconstruction. 
\end{IEEEbiography}
\begin{IEEEbiography}
[{\includegraphics[width=1.15in,height=1.35in,clip,keepaspectratio]
{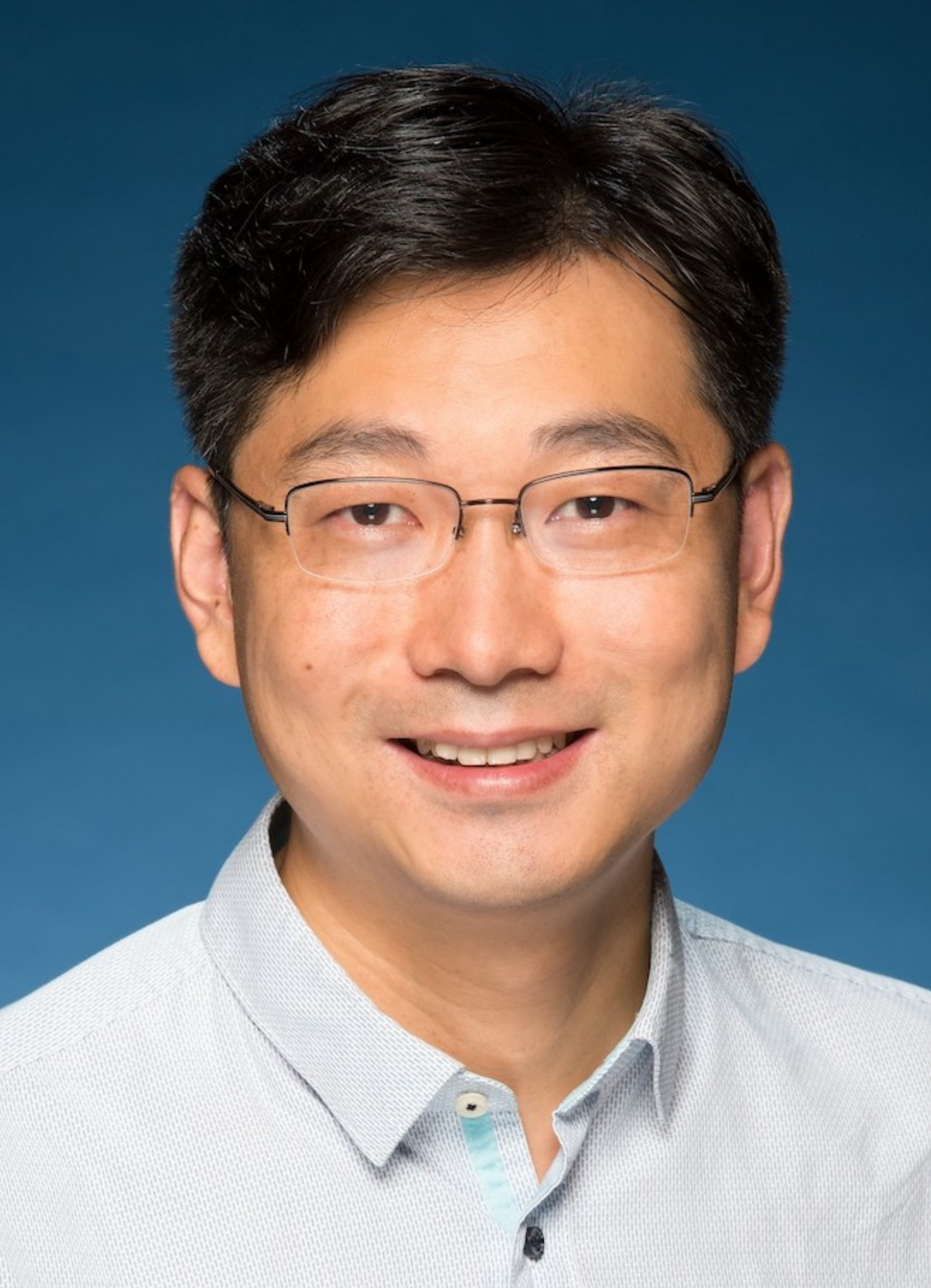}}]{Xiaowen Chu}
received the B.E. degree in computer science from Tsinghua University, P.R. China, in 1999, and the Ph.D. degree in computer science from The Hong Kong University of Science and Technology in 2003. Currently, he is a full professor in the Department of Computer Science, Hong Kong Baptist University. His research interests include distributed and parallel computing and wireless networks. He is serving as an Associate Editor of IEEE Access and IEEE Internet of Things Journal.
\end{IEEEbiography}

% You can push biographies down or up by placing
% a \vfill before or after them. The appropriate
% use of \vfill depends on what kind of text is
% on the last page and whether or not the columns
% are being equalized.

%\vfill

% Can be used to pull up biographies so that the bottom of the last one
% is flush with the other column.
%\enlargethispage{-5in}

% that's all folks
\end{document}